\pdfoutput=1
\documentclass{article}

\usepackage[dvipsnames,table]{xcolor}
\usepackage[utf8]{inputenc}
\usepackage[T1]{fontenc}

\usepackage[preprint]{corl_2026} 

\usepackage{amsmath}
\usepackage{amsfonts}
\usepackage{threeparttable}
\usepackage{booktabs}
\usepackage{graphicx}
\usepackage{etoolbox}
\usepackage{subcaption}
\usepackage{multirow}
\usepackage{siunitx}
\usepackage{caption}
\usepackage{pifont}
\newcommand{\cmark}{\ding{51}}
\usepackage{float}
\usepackage{wrapfig}

\newcommand{\waglong}{Wiggle and Go!}
\newcommand{\wag}{WaG}

\title{\waglong{} System Identification for Zero-Shot Dynamic Rope Manipulation}

\author{
  Arthur Jakobsson$^{1}$,
  Abhinav Mahajan$^{1}$,
  Karthik Pullalarevu$^{1}$,
  Krishna Suresh$^{1}$,
  Yunchao Yao$^{2}$ \\
  Yuemin Mao$^{1}$,
  Bardienus Duisterhof$^{1}$,
  Shahram Najam Syed$^{1}$,
  Jeffrey Ichnowski$^{1}$ \\[0.6em]
    \mdseries\normalsize $^{1}$Robotics Institute, Carnegie Mellon University \\
    \mdseries\normalsize $^{2}$University of North Carolina at Chapel Hill
}

\begin{document}
\maketitle

\begin{figure}[H]
    \centering
    \includegraphics[width=1\textwidth]{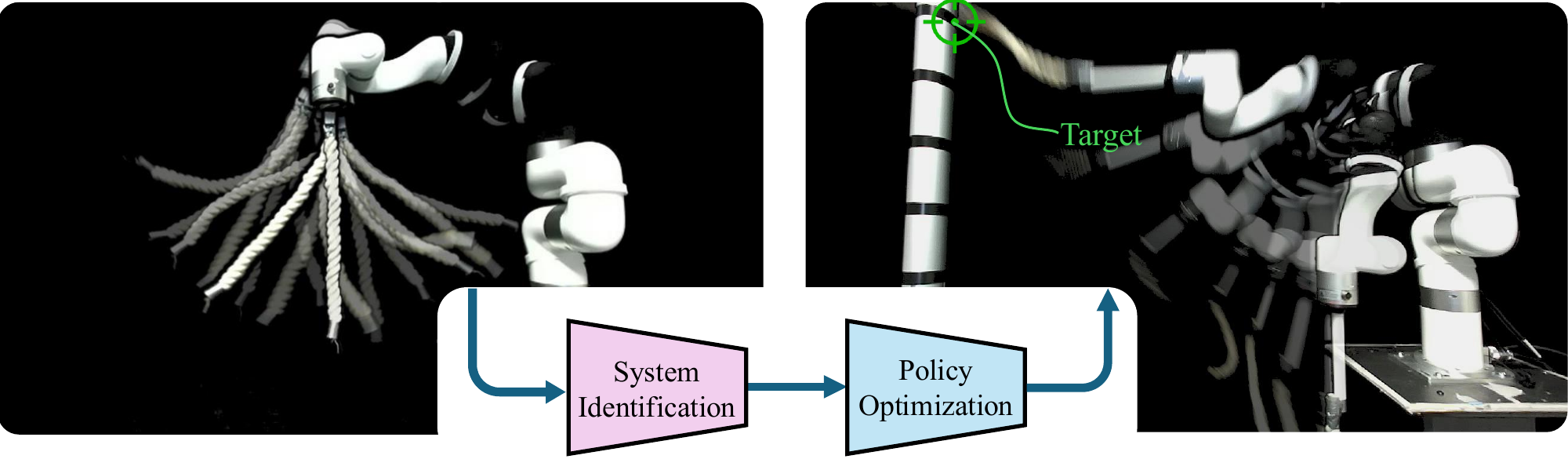}
    \captionof{figure}{\waglong{} uses a brief \textit{wiggle} motion (left) to identify the dynamic behavior of a rope using a neural network model.
    This is fed into a goal-conditioned policy optimizer, which then \textit{goes!}, performing a dynamic fling (right), hitting the target in one shot. Our system identification model and policy optimizer work entirely in simulation enabling the wiggle and task rollout to be the only actions we perform in real.}
    \label{fig:system_overview}
\end{figure}


\begin{abstract}

Many robotic tasks are unforgiving; a single mistake in a dynamic throw can lead to unacceptable delays or unrecoverable failure. We introduce Wiggle and Go!, a two-stage framework for zero-shot rope manipulation: a brief, safe \emph{wiggle} action is observed to predict descriptive rope parameters, which then conditions a trajectory optimizer for zero-shot goal-conditioned execution. Unlike prior dynamic rope manipulation methods that require large real-world datasets or iterative real-world refinement, our identification module is task-agnostic, supporting diverse manipulation policies without retraining. We achieve a 3.55\,cm average accuracy on 3D target striking in real using rope system parameters in comparison to 15.29\,cm for uninformed baselines, and over 50\% success on multi-objective lobbing and draping tasks. Predicted parameters transfer to unseen motions with 0.95 Pearson correlation between simulated and real rope dynamics, indicating that the identification module generalizes across the task corpus.
Project website: \normalsize\url{https://wiggleandgo.github.io/}
\end{abstract}




\section{Introduction}

Before a sailor throws a mooring line or a climber flings a rope, they might instinctively give it a brief shake. This `wiggle' is not just a habit; it is a sophisticated probe: the human nervous system reads the rope's characteristics to calibrate the subsequent throw. Robots performing similar dynamic tasks are blind to these hidden dynamics, and when failures are unrecoverable (tangled, broken, or caught ropes) or costly, iterating over attempts is too risky.

Dynamic rope manipulation is hard precisely because rope behavior emerges from an interplay of stiffness, damping, inertia, and geometry that varies across ropes and cannot be read from a label. Iterative refinement methods~\citet{chi_iterative_2022} and~\citet{zhang_robots_2020} achieve strong results but require 5-10 real-world trials \emph{per goal}, a cost that is unacceptable when a single failed throw risks equipment damage or an unrecoverable tangle. Large-dataset approaches~\cite{nair_combining_2017} require tens of thousands of autonomous interactions and do not adapt to a new rope without re-collection. Parameter-conditioned methods,~\citet{kuroki_genorm_2025, kuroki_gendom_2025}, estimate material properties such as Young's modulus from point-cloud observations, but these do not map one-to-one onto the behavioral parameters needed for physics-based trajectory optimization of dynamic 3D tasks.

We bridge this gap by replicating the sailor's strategy: a single short, safe probing action decodes a rope's physical identity before committing to a high-stakes dynamic maneuver. We propose \waglong{} (\wag{}), a two-stage framework in which (a) a brief \emph{wiggle} excites rope dynamics for task-agnostic system identification, and (b) the predicted parameters condition a zero-shot trajectory optimizer. Because identification is decoupled from execution, one wiggle supports diverse downstream tasks without retraining. Our primary contributions are: a task-agnostic system identification module, trained entirely in simulation, that infers nine behavioral rope parameters from a single planar wiggle and transfers to unseen motions at 0.95 Pearson correlation between simulated and real rope dynamics; a decoupled two-stage pipeline in which one wiggle observation supports zero-shot deployment across qualitatively different tasks (striking, lobbing, draping) without task-specific retraining or additional real-world trials; and a comprehensive real-world evaluation (with over 600 trials) across diverse ropes that achieves a four-fold accuracy improvement on 3D target striking over our baseline (\textbf{3.55\,cm} vs.\ 15.29\,cm) and over 50\% success on lobbing and draping, with a parameter-importance ablation establishing that non-directly-measurable properties (stiffness, damping, mass distribution) drive ${\sim}10$\,cm of the improvement.

\section{Related Works}
\label{sec:citations}

\noindent\textbf{Rope Dynamics Through Data and Simulation.} Learning rope dynamics from large-scale data is constrained by the difficulty and safety of collecting diverse real-world demonstrations. Early work~\cite{nair_combining_2017} required approximately 60{,}000 autonomous interactions; self-supervised methods~\cite{wang2024selfsupervisedlearningdynamicplanar} reduce this cost but still depend on extensive rollouts. Benchmark and imitation-learning approaches~\cite{lin2020softgym, zhao2023learning, luo2024multistage} scale via demonstrations rather than environment interaction, but neither is well-suited to dynamic settings where individual failures can damage equipment or cause safety hazards.

Sim-to-real transfer offers an alternative by training primarily in simulation. Domain randomization methods~\cite{matas_simreal_2018, tiboni2023dropo} train policies across varied simulation parameters to transfer policies to real deformable objects. Differentiable physics simulators~\cite{hu2019chainqueen, jatavallabhula2021gradsim, chen2024deform, si2024difftactile} enable gradient-based parameter identification and trajectory optimization. Trajectory-optimization approaches for deformable linear objects~\cite{jnadi2026scope, artinian2024optimal, yamakawa_simple_2012, yamakawa_dynamic_2013} synthesize motions with explicit physics constraints, spanning quasi-static Cosserat-rod shape control to analytical high-speed models for specific dynamic tasks. Real-time cable-shaping MPC~\cite{viljoen2024cable} similarly operates in quasi-static regimes due to re-planning cost. We show that identification from a wiggle complements these approaches by supplying better parameter estimates for sub-second dynamic motions where pre-execution planning suffices.

\noindent\textbf{System Identification for Deformable Objects.} System identification approaches explicitly estimate physical parameters to bridge simulation and reality. Real2Sim2Real~(\citet{lim_planar_2022}) optimizes simulator parameters to match observed trajectories for dynamic casting, though requiring extensive physical experiments for each new cable. GenORM~(\citet{kuroki_genorm_2025}) and GenDOM~(\citet{kuroki_gendom_2025}) condition policies on Young's modulus and Poisson's ratio estimated from point cloud observations during predefined lifting motions, demonstrating one-shot manipulation on single-stage planar tasks. \waglong{} differs in three key ways: (1) we use higher-dimensional, rope-specific parameters rather than material properties alone, \citet{lim_planar_2022} argues that Young's modulus-based simulation can create discrepancies for ropes with leads, (2) we demonstrate transferability across multiple complex 3D manipulation tasks rather than single-stage goals, and (3) we perform an analysis of how observation strategy (wiggle design) affects parameter prediction quality.

Learning-based dynamics models~\cite{yang2022learning, wang2022offline, yu_shape_2022} capture rope behavior through learned recurrent, graph, or local-linear representations; these predict implicit state-space rollouts rather than interpretable physical parameters, so they cannot be instantiated inside an analytic physics simulator for parameter-based trajectory optimization. Unlike privileged-learning frameworks such as Rapid Motor Adaptation~\cite{kumar_rma_2021, qi_-hand_2022, liang_rapid_2024} that infer hidden factors implicitly during execution, we estimate parameters explicitly and offline through a safe observation action, enabling task-agnostic identification. Concurrent work,~\citet{sobanbabu2025spi}, also uses CMA-ES--based sim-ID, but in legged locomotion and via iterative active exploration; we instead identify a rope from a single wiggle for one-shot, zero-shot deployment.

\begin{figure*}[t]
    \centering
    \includegraphics[width=1\textwidth]{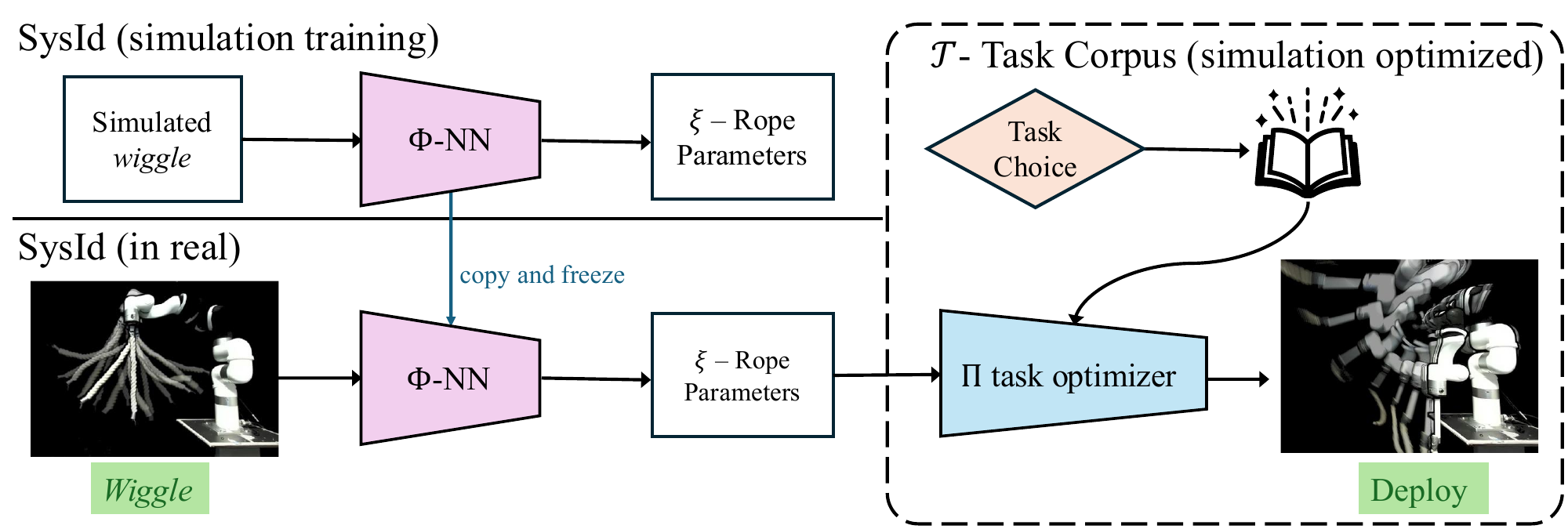}
    \caption{The \wag{} pipeline. $\Phi$-NN is trained entirely in simulation; at deployment, a real-world wiggle feeds $\Phi$-NN, whose predicted parameters inform a task-specific trajectory optimizer that is then executed on the robot. Items in \textcolor{Green}{green} are the only real-world steps, executed once each.}
    \label{fig:pipeline}
\end{figure*}

\noindent\textbf{Iterative Learning Methods.} Iterative learning control refines policies through repeated trials~\citet{bristow_survey_2006}. For dynamic rope manipulation, \citet{chi_iterative_2022, zhang_robots_2020} achieve strong results across diverse ropes and tasks through progressive refinement, typically requiring 5-10 trials per goal to achieve precise manipulation. \citet{suresh2026learning} extend dynamic rope manipulation task-level ILC, learning a flying-knot skill with a critical-point objective on the rope-state trajectory. Our method builds towards addressing scenarios where repeated attempts pose risks of rope damage, equipment failure, or environmental hazards. Critically different, our system identification module is task-agnostic so the same wiggle observation can support multiple downstream manipulation tasks, enabling seamless switching between different goals without additional real-world trials.

\section{Method}

\waglong{} has two stages: a safe \emph{wiggle} excites rope dynamics for parameter estimation, and the predicted parameters condition a task-specific trajectory optimizer. Decoupling observation from execution lets one wiggle support multiple downstream tasks without retraining. Formally, \wag{} observes rope motion during the wiggle to obtain visual features $\mathcal{O}$.
The System Identification (SysId) model $\Phi: \mathcal{O} \rightarrow \Xi$ predicts rope system parameters $\xi \in \Xi$ (e.g., stiffness, damping, mass distribution, link count). Given a task goal $g \in \mathcal{G}$ (e.g., 3D target position) and rope parameters $\xi$, an action policy $\Pi: \Xi \times \mathcal{G} \rightarrow \mathcal{A}$ generates a robot trajectory $\vec{a} \in \mathcal{A}$.

A task corpus $\mathcal{T}$ collects per-task policies (target striking, lobbing, draping). Because $\Phi$ is task-agnostic, every policy in $\mathcal{T}$ takes the same input form.
Figure~\ref{fig:pipeline} illustrates the complete system architecture.

\subsection{Action Policy Optimization $\Pi$} \label{sec:cmaes-traj}

Each task in the corpus requires a policy $\Pi_{\text{task}}: \Xi \times \mathcal{G} \rightarrow \mathcal{A}$ that maps rope parameters and goals to executable trajectories. 
To avoid real-world trial and error, we optimize trajectories in simulation using predicted rope parameters $\hat{\xi}$ from the adaptation model~(\ref{phi_NN}). 

We use Drake~\cite{drake} for high-fidelity rope simulation of dynamic motions.
We employ data-efficient, derivative-free optimization that can discover effective trajectories within a few hundred simulations per rope-task pair.

\begin{wrapfigure}{r}{0.5\textwidth}
    \vspace{-0.25em}
    \centering
    \includegraphics[width=0.5\textwidth]{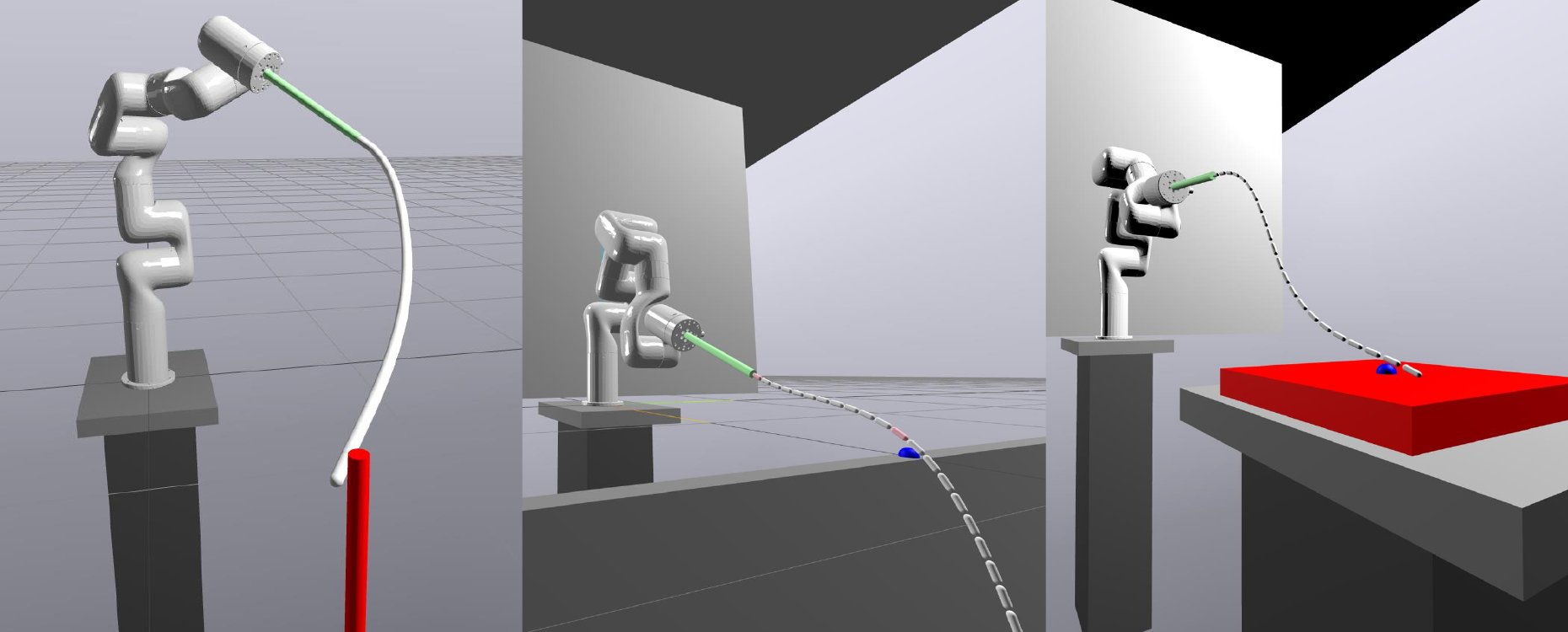}
    \caption{Drake tasks left to right: 3D striking, draping, lobbing. \textcolor{SeaGreen}{Green:} extender; \textcolor{RedOrange}{Red}/\textcolor{Blue}{Blue:} targets}
    \label{fig:drake_sim}
    \vspace{-0.25em}
\end{wrapfigure}

\noindent\textbf{Task Setup.} We choose three toy tasks to represent our task corpus: 3D point striking, lobbing, and draping shown in Figure~\ref{fig:drake_sim}. The lobbing task is motivated by controlled placement strategies in deformable object manipulation~\cite{chen2022garment}. Lobbing and draping use multi-objective reward formulations defined in Appendix~\ref{app:rewards}. Our task setup (Figure~\ref{fig:drake_sim}) uses a UFactory xArm 7 robot manipulator~\cite{ufactory2022xarm7}. 
For all tasks, we attach ropes to a 20\,cm pole extender to increase the moment arm; the xArm's joint velocity limits require this leverage for dynamic motions like whipping. The optimization objective minimizes 3D Euclidean distance between the rope tip and target.

\noindent\textbf{CMA-ES Trajectory Optimization.}\label{sec:cmaes_traj} We use Covariance Matrix Adaptation Evolution Strategy (CMA-ES)~\cite{cmaes}, a stochastic optimization algorithm that adapts its search distribution based on the success of sampled solutions. At each iteration, CMA-ES-traj (CMA-ES for trajectory optimization) samples a population of candidate trajectories from a multivariate Gaussian distribution, evaluates their fitness (task error in simulation), and updates the distribution's mean and covariance matrix to concentrate search around promising regions. Each task uses a task-specific fitness combining a distance term with task-shaped auxiliaries: a proximity reward and motion regularizer for pole striking, an additional dwell term for lobbing that activates once the tip enters the target neighborhood, and an intermediate waypoint plus height penalty for draping. Joint-velocity and collision penalties are folded into the fitness directly. Formal definitions are given in Appendix~\ref{app:rewards}. \

We parameterize trajectories using three sequential waypoints in the robot's joint space and interpolate with a cubic spline to generate smooth, continuous joint trajectories within velocity limits. This low-dimensional representation (21D for a 7-DOF arm) is substantially smaller than the full per-step joint trajectory. For the more complex tasks, we limit exploration to planar actions by optimizing only the three joints that move in the plane of motion (9D parameters). Joint-space waypoints also avoid inverse kinematics singularities and naturally respect joint limits and velocity constraints. For each rope parameter set $\xi$ and goal $g$, CMA-ES-traj runs for up to 25 iterations with 60 samples per iteration. The initial sampling distribution is unbiased for 3D pole-striking; for lobbing and draping, where the optimization landscape is more multi-modal, we bias the initial distribution toward a lifting motion as a warm start.

\noindent\textbf{Simulation Setup and Implementation.} We represent ropes in simulation as link/ball-joint chains, following Lim et al.~\cite{lim_planar_2022}, where the rope is modeled as a series of rigid spheres connected by ball joints with stiffness $k$ and damping parameters $c$. Each link has a mass and radius for accurate inertial calculations. We train on parameters spanning the ranges in Table~\ref{tab:in_domain_results}, chosen to match real ropes in household and industrial settings.
We also, following \cite{lim_planar_2022}, include lead weights at the rope tip to model common real-world scenarios where ropes are used to cast or manipulate attached objects. The lead weight also increases the effective reach by providing momentum to the rope tip during dynamic motions. The nine behavioral descriptors in Table~\ref{tab:in_domain_results} characterize the rope's dynamic response inside our simulator and do not correspond one-to-one to measurable physical properties.

\subsection{Adaptation Model $\Phi$}\label{phi_NN}

Trajectory optimization (Section~\ref{sec:cmaes-traj}) requires the rope's parameters. Even when physical properties are available, they do not map one-to-one onto the behavioral parameters of our ball-joint simulator, so we infer them from observed simulation behavior. The mapping from motion to parameters is non-linear; we frame parameter inference as supervised learning and train $\Phi$-NN ($\Phi$-neural network) on simulated ropes that span the ranges in Table~\ref{tab:in_domain_results}, each with known ground-truth labels that are unavailable for physical ropes (e.g.\ stiffness, damping, and link count cannot be directly measured). At deployment, $\Phi$-NN consumes a wiggle observation $\mathcal{O}$ and returns $\hat{\xi}$, which is then passed to the trajectory optimizer.

\subsubsection{Observation Collection with Safe Action}\label{sec:howtowiggle}

We collect observations by executing a planar wiggle trajectory $\omega_{\text{wiggle}}$ that oscillates the rope with the robot end-effector. The wiggle is designed to excite key rope properties (stiffness, damping, inertia, etc.) through controlled acceleration while maintaining visibility for tracking. An ablation in the appendix (Sec.~\ref{sec:wiggleablation_appendix}) shows that the specific wiggle trajectory is not critical; any motion that sufficiently excites rope dynamics performs comparably.\footnote{We do not provide any theoretical guarantees on the safety of the wiggle action, but we use this terminology to suggest that a known, more controlled action is less likely to cause physical damage to surroundings and itself.}

In simulation, for a rope with $N$ links, we extract 3D positions of the link centers $\{p_t^{(i)} \in \mathbb{R}^3\}_{i=1}^N$ at each timestep $t$. We project these to 2D image coordinates using a calibrated pinhole camera model. Since our wiggle is planar by design, depth information is largely redundant, the rope motion occurs primarily in the image plane and all links remain visible from a single fixed viewpoint. While stereo setups with non-planar motion could potentially improve inference of damping and stiffness, the single-camera approach balances simplicity with empirically validated performance (Section~\ref{sec:fullpipeline}).

\textbf{Feature Engineering.} We normalize link positions at deployment relative to the first link and use only position and angle features, $\mathcal{O} = \{\tilde{p}_t, \theta_t\}$, omitting velocity and acceleration of the points to ensure minimal feature drift during real-world deployment. Details are in Appendix~\ref{sec:features_appendix}.

\subsubsection{Network Architecture and Sim-to-Real Training}

$\Phi$-NN is a temporal convolutional encoder followed by an MLP that maps observations to normalized parameters $\hat{\xi}\in[0,1]^9$ (Table~\ref{tab:in_domain_results}). We close the sim-to-real gap through domain randomization (calibration noise, anisotropic temporally-correlated tracking noise, and randomized trajectory padding) and curriculum masking that progressively occludes random and beginning-biased frame blocks to prevent overfitting to specific temporal windows. Full architecture, randomization, and training specifications are in Appendix~\ref{sec:imp_details}.

\subsection{Real-World Deployment}
Real-world deployment uses a ZED Mini 2i camera calibrated to the simulation's pinhole model; Grounding SAM~\cite{ren2024grounded} segments the rope in the first frame, and CoTracker~\cite{karaev2410cotracker3} tracks centerline keypoints across the wiggle. The tracked 2D keypoints are converted to the same normalized position-and-angle features used during training and passed to $\Phi$-NN to obtain $\hat{\xi}$; CMA-ES-traj is then run offline in Drake with $\hat{\xi}$, and the resulting joint trajectory is executed open-loop on the xArm 7. No real-world feedback is used during task execution, so residual real-world error reflects sim-to-real gap and identification quality rather than online correction. Task error is measured by HSV-based endpoint tracking, with depth from the ZED for 3D distance. Full deployment details are in Appendix~\ref{sec:deployment_appendix}.

\section{Experiments and Results}

We perform a series of experiments to evaluate the efficacy of our adaptation model, the transferability of rope system parameters from one dynamic context to another, and the full-pipeline accuracy of our system-identification-enabled zero-shot manipulation method. We perform our tests on a diverse range of ropes and a chain. The chain represents an out-of-distribution example of a \textit{rope-behaving} object, though we note that ropes are more accurately represented as ball joints than chains which would be closer to an alternating hinge joint.

\begin{figure}[t]
\centering
\begin{minipage}[c]{0.45\linewidth}
\centering
\includegraphics[width=\linewidth]{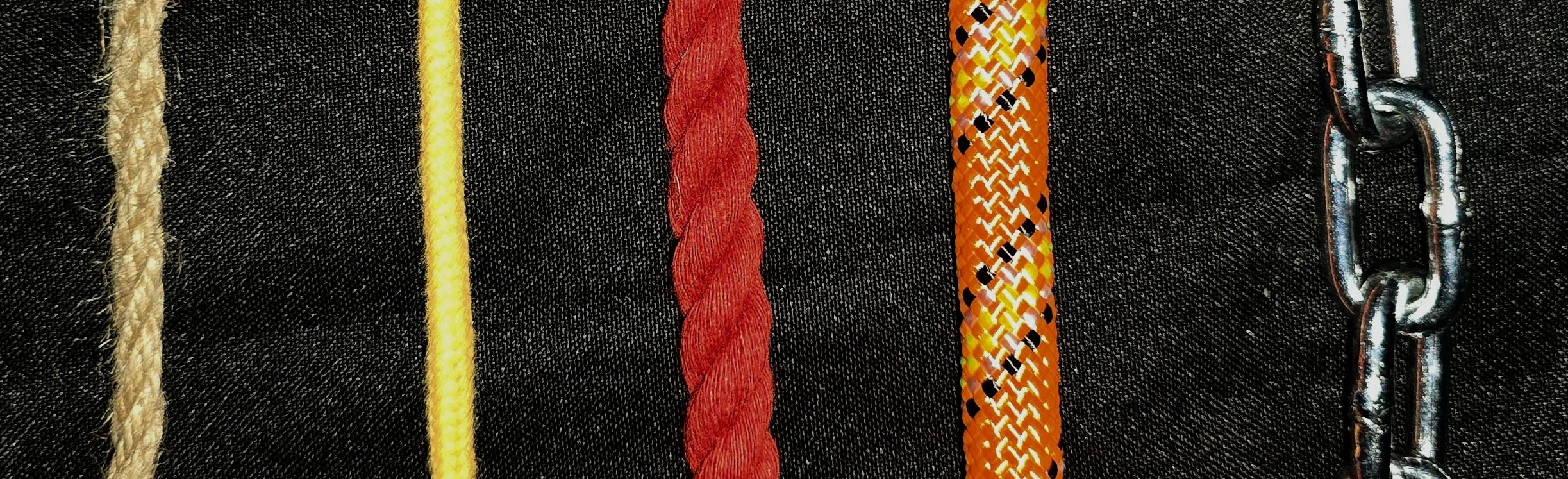}
\caption{Test ropes (order: \textcolor{Brown}{Brown}, \textcolor{Dandelion}{Yellow}, \textcolor{Red}{Red}, \textcolor{Orange}{Orange}, \textcolor{CadetBlue}{Chain}).}
\label{fig:ropes}
\end{minipage}
\hfill
\begin{minipage}[c]{0.50\linewidth}
\centering
\small
\setlength{\tabcolsep}{3pt}
\captionof{table}{Material and physical characterization of test ropes.}
\label{tab:rope_descriptions}
\resizebox{\linewidth}{!}{%
\begin{tabular}{lccccc}
\toprule
\textbf{Rope} & \textbf{Mat.} & \textbf{Stiff.} & \textbf{Damp.} & \textbf{Wt.} & \textbf{Rad.} \\
& & & & \textbf{(kg/m)} & \textbf{(mm)} \\
\midrule
\textcolor{Brown}{Brown} & Twine & Slack & Low & 0.013 & 6 \\
\textcolor{Dandelion}{Yellow} & Cotton & Stiff & Med & 0.011 & 5 \\
\textcolor{Red}{Red} & Cotton & Slack & Low & 0.033 & 9.5\\
\textcolor{Orange}{Orange} & Polyester & Stiff & High & 0.082 & 12.7\\
\textcolor{CadetBlue}{Chain} & Steel & Slack & Low & 0.52 & 4.8$\times$18\\
\bottomrule
\end{tabular}}
\end{minipage}
\end{figure}

The five ropes are shown in Fig.~\ref{fig:ropes} and characterized in Tab.~\ref{tab:rope_descriptions}. They span the qualitative axes of the training distribution: slack/stiff stiffness, low/medium/high damping, 0.011--0.082\,kg/m, and 5--12.7\,mm radius (from thin twine to stiff polyester). We evaluate the information conveyed by the wiggle, the simulation precision, the cross-motion transferability of predicted parameters, and the importance of identification for real-world rollout.

\subsection{Adaptation Model Evaluation}

We evaluate $\Phi$-NN on three criteria: parameter estimation accuracy, robustness to distribution shift, and transferability to downstream tasks. All experiments use the model trained on 9000 simulated ropes with domain randomization enabled.

\subsubsection{Parameter Estimation Accuracy and Robustness}

\begin{wraptable}{r}{0.45\textwidth}
\vspace{-0.5em}
\centering
\small
\caption{In-distribution parameter ranges and mean absolute error on 1000 test samples.}
\label{tab:in_domain_results}
\resizebox{\linewidth}{!}{%
\begin{tabular}{lccc|c}
\toprule
& \multicolumn{3}{c|}{\textbf{Parameter Range}} & \\
\cmidrule(lr){2-4}
\textbf{Parameter} & \textbf{Min} & \textbf{Max} & \textbf{Units} & \textbf{MAE} \\
\midrule
Number of links & 20 & 26 & --- & 0.098 \\
Rope length & 0.45 & 0.65 & m & 0.006 \\
Ball damping & 0.001 & 0.05 & N$\cdot$s/m & 0.010 \\
Ball stiffness & 0.05 & 1.0 & N/m & 0.111 \\
Rope radius & 0.003 & 0.015 & m & 0.002 \\
Mass per unit len & 0.02 & 0.12 & kg/m & 0.007 \\
Lead mass & 0.02 & 0.12 & kg & 0.005 \\
Lead radius & 0.015 & 0.045 & m & 0.008 \\
Link extra scale & 50 & 120 & \% & 0.152 \\
\midrule
\multicolumn{4}{l|}{\textbf{Overall}} & \textbf{0.045} \\
\bottomrule
\end{tabular}}
\end{wraptable}

On a test set of 1000 held-out simulated ropes, $\Phi$-NN reaches sub-millimeter MAE on geometric parameters (rope length 6\,mm, rope radius 2\,mm, number of links 0.098). Inertial and damping parameters have larger absolute MAE (ball damping 0.010\,N$\cdot$s/m, ball stiffness 0.111\,N/m, mass per unit length 0.007\,kg/m), reflecting the difficulty of inferring these from a planar projected motion. Parameter Importance (Section~\ref{sec:param_importance}) demonstrates that these absolute errors still leave the predictions useful for downstream trajectory optimization, and multicollinearity in the ball-joint model allows neighboring parameters to compensate for one another. We additionally find that the choice of wiggle is not critical: alternative planar wiggles of varied amplitude and frequency yield comparable parameter estimates (Appendix~\ref{sec:wiggleablation_appendix}), and the model gracefully clamps to its training bounds rather than extrapolating on out-of-distribution parameters (Appendix~\ref{sec:ood_appendix}).

\subsubsection{Transferability to Downstream Tasks}

Parameter-wise accuracy alone is insufficient and predicted parameters must work together to enable task execution. We therefore validate transferability through motion fidelity (whether the rope with predicted parameters reproduces real rope dynamics under different motions) and task-level performance (whether they enable effective trajectory optimization).

\textbf{Baseline: $\Phi$-CMA-ES.} To benchmark $\Phi$-NN, we implement an optimization-based baseline that directly fits simulation parameters to a real wiggle trajectory: 60 samples per iteration $\times$ 50 CMA-ES iterations minimize pixel-wise distance between the simulated and real 2D rope trajectories. Without learned priors, $\Phi$-CMA-ES intentionally overfits each rope to its wiggle, costing $\sim$3000 simulations per inference but offering robustness to out-of-distribution objects like chains where $\Phi$-NN's priors no longer apply.

\begin{figure}[t]
    \centering
    \includegraphics[width=0.48\textwidth]{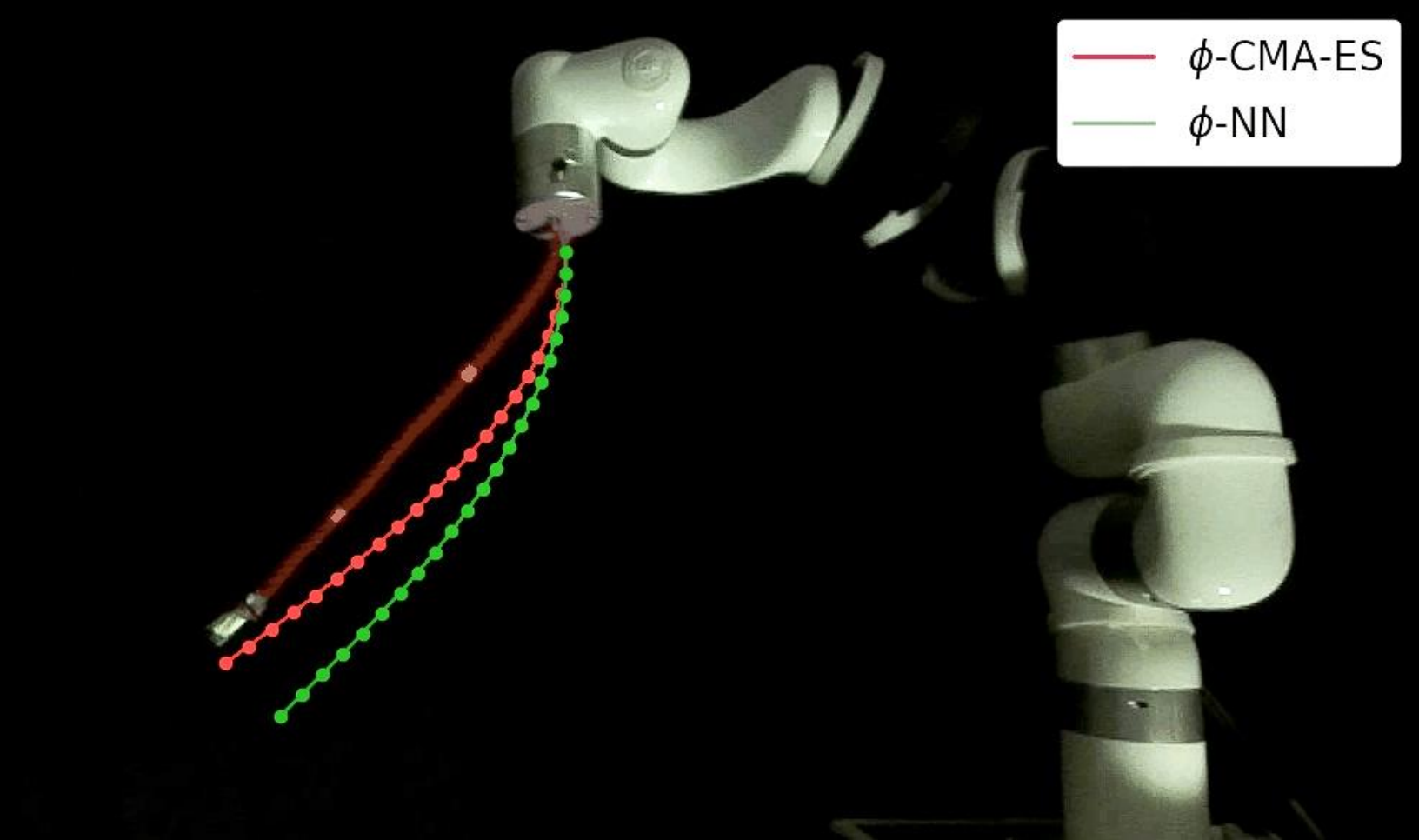}\hfill
    \includegraphics[width=0.48\textwidth]{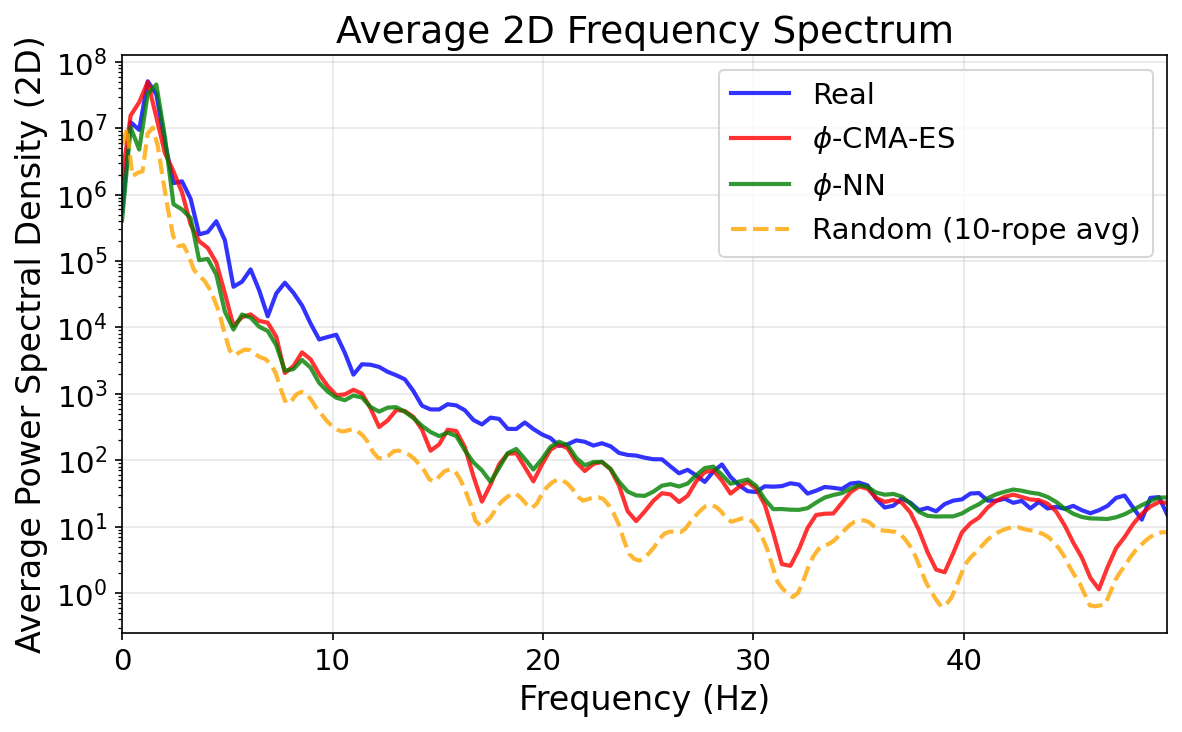}
    \caption{Real vs.\ simulation comparison for $\Phi$-NN and $\Phi$-CMA-ES on an unseen dynamic motion (Red rope, 45\,cm, 10\,g lead): point-wise trajectory overlay (left) and Fourier frequency distribution (right). See Appendix~\ref{sec:tranferability-appendix} for similar trends across ropes.}
    \label{fig:motion_fidelity}
\end{figure}

\textbf{Motion Fidelity.} To test fidelity beyond static accuracy, for 10 real ropes (5 types $\times$ 10/30\,g leads) we predict parameters from one wiggle, then compare a held-out motion in sim and real via Fourier and point-wise trajectory analysis (Fig.~\ref{fig:motion_fidelity} and details in Appendix~\ref{sec:tranferability-appendix}). Both $\Phi$-NN and $\Phi$-CMA-ES achieve 0.95 average frequency correlation. $\Phi$-NN achieves 5.4\,cm/frame average error versus 5.8\,cm/frame for $\Phi$-CMA-ES, both substantially outperforming random parameters (13.6\,cm/frame). The ${\sim}2.5\times$ gap over random parameters indicates the wiggle-inferred parameters reproduce rope dynamics throughout an unseen trajectory. $\Phi$-CMA-ES does better on the out-of-distribution chain (0.99 vs.\ 0.78 correlation), where per-rope fitting beats learned priors.

\textbf{Task-Level Validation.} We first validate in simulation that predicted parameters enable downstream trajectory optimization. For 100 random simulated ropes, we predict parameters with $\Phi$-NN, optimize trajectories using those predictions (CMA-ES-traj, Section~\ref{sec:cmaes-traj}), and measure task error (Table~\ref{tab:sim-full}). Trajectories optimized with predicted parameters achieve 2.1\,cm median error versus 1.2\,cm with ground truth and 12.8\,cm with random in-distribution parameters. Predicted parameters provide substantial improvement over uninformed baselines while remaining within 0.9\,cm of ground truth performance. This demonstrates that parameter estimation errors do not compound catastrophically, suggesting $\Phi$-NN provides sufficient information for control, validating the two-stage pipeline design.

\subsection{Action Policy Optimization Evaluation} \label{sec:cma-es-traj-eval}

To attribute residual real-world error to SysId quality rather than the optimizer, we need an optimizer that reliably finds near-optimal trajectories given ground-truth parameters. CMA-ES-traj satisfies this: across 100 ground-truth ropes (Appendix Tab.~\ref{tab:sim-full}) it reaches 1.2\,cm median and 0.25\,cm minimum tip-to-target error. We pre-filter samples for joint-velocity feasibility so reported errors come from hardware-feasible trajectories. The 25--120 minute CPU cost per rope-task is acceptable when each rope is identified once and keeps \textit{system identification quality} as the primary varying factor across our experiments. An RL policy conditioned on $\hat{\xi}$ is a viable alternative that could amortize this cost at deployment; it would, however, require additional reward, exploration, and architecture tuning and may converge less uniformly across the rope range than the per-task optimization used here. The higher convergence times were largely due to simulations with more rope-object contacts.

Across tasks CMA-ES-traj converges consistently within budget (25 iterations for pole striking, 50 for lobbing, 15 for draping). On pole striking it reaches 0.8\,cm average tip-to-target distance; with a simulation timestep of 0.01\,s and the high speeds of these motions, the ground-truth instant of minimum distance is often missed by a small margin, so the true achieved accuracy is likely lower than reported.

\subsection{Full Pipeline Results in Real} \label{sec:fullpipeline}

We test four in-domain ropes (Fig.~\ref{fig:ropes}) at 45/55/65\,cm with 5\,g lead and the 45\,cm rope with 10/20/30\,g leads; a metal chain serves to demonstrate capabilities of out-of-distribution generalization. The benchmark $\Phi$-Random executes trajectories optimized for 20 unrelated in-distribution ropes, simulating an educated but unmatched action (omitted on the chain for safety). $\Phi$-NN outperforms $\Phi$-CMA-ES on in-distribution ropes (3.55\,cm vs.\ 4.16\,cm average), and $\Phi$-CMA-ES wins on the chain by overfitting to its dynamics.

\textbf{Parameter Importance: Do Other Parameters Actually Move the Outcome?}
\label{sec:param_importance}
A reasonable concern is that the two directly-measurable parameters (length, lead mass) dominate the benefit and the other seven are cosmetic. Replacing subsets of $\Phi$-NN's predictions with random in-distribution values (Appendix Tab.~\ref{tab:param_importance}) shows otherwise: full identification reaches 0.9\,cm mean error versus 14.2\,cm for fully random, but retaining only length and lead-mass closes just 3.3\,cm of that 13.3\,cm gap: the remaining seven parameters (stiffness, damping, mass distribution) drive ${\sim}10$\,cm. The real-world data in Table~\ref{tab:rope_results} agrees: $\Phi$-NN reduces mean error over $\Phi$-Random by 3.3--13.1$\times$, and ropes sharing length and lead but differing in stiffness/damping span a 7.5\,cm range in $\Phi$-Random error.

\begin{table}[t]
\footnotesize
\centering
\caption{Full pipeline results after rope wiggle, parameter prediction and CMA-ES-traj execution. Average distance (cm) from target from 600 task-policy rollouts (5 per rope-config-method). Each (length, lead) configuration shows three methods side-by-side: \textbf{NN}~=~$\Phi$-NN, \textbf{CMA}~=~$\Phi$-CMA-ES, \textbf{Rnd}~=~$\Phi$-Random. Best in-domain method per (rope, config) in \textbf{bold}.}
\label{tab:rope_results}
\setlength{\tabcolsep}{3pt}
\resizebox{\textwidth}{!}{%
\begin{threeparttable}
\begin{tabular}{l ccc ccc ccc ccc ccc ccc ccc}
\toprule
 & \multicolumn{3}{c}{\textit{45cm, 5g}} & \multicolumn{3}{c}{\textit{45cm, 10g}} & \multicolumn{3}{c}{\textit{45cm, 20g}} & \multicolumn{3}{c}{\textit{45cm, 30g}} & \multicolumn{3}{c}{\textit{55cm, 5g}} & \multicolumn{3}{c}{\textit{65cm, 5g}} & \multicolumn{3}{c}{\textbf{Average}} \\
\cmidrule(lr){2-4} \cmidrule(lr){5-7} \cmidrule(lr){8-10} \cmidrule(lr){11-13} \cmidrule(lr){14-16} \cmidrule(lr){17-19} \cmidrule(lr){20-22}
\textbf{Rope} & NN & CMA & Rnd & NN & CMA & Rnd & NN & CMA & Rnd & NN & CMA & Rnd & NN & CMA & Rnd & NN & CMA & Rnd & NN & CMA & Rnd \\
\midrule
\textcolor{Brown}{Brown}      & 0.1          & \textbf{0.0} & 15.9 & 2.0          & \textbf{0.6} & 17.3 & \textbf{5.3} & 8.9          & 18.3 & \textbf{0.8} & 5.2          & 17.7 & \textbf{0.0} & 0.2          & 15.5 & \textbf{2.1} & 4.4          & 22.1 & \textbf{1.7} & 3.2          & 17.8 \\
\textcolor{Dandelion}{Yellow} & 8.1          & \textbf{1.5} & 14.7 & 6.3          & \textbf{3.8} & 15.8 & 0.8          & \textbf{0.8} & 16.5 & 4.7          & \textbf{3.2} & 17.0 & \textbf{3.5} & 7.7          & 15.0 & \textbf{0.0} & 8.1          & 19.2 & \textbf{3.9} & 4.2          & 16.4 \\
\textcolor{Red}{Red}          & \textbf{0.0} & \textbf{0.0} & 17.6 & 16.0         & \textbf{3.2} & 15.2 & \textbf{0.9} & 2.3          & 16.9 & \textbf{1.4} & 7.2          & 16.6 & \textbf{0.3} & 1.4          & 14.1 & 4.3          & \textbf{0.0} & 18.2 & 3.8          & \textbf{2.4} & 16.4 \\
\textcolor{Orange}{Orange}    & 10.1         & \textbf{0.0} & 10.9 & \textbf{2.2} & 8.7          & 10.0  & 3.7          & \textbf{1.5} & 11.8 & \textbf{0.0} & \textbf{0.0} & 13.5 & \textbf{11.4}& 12.0         & 8.0  & \textbf{1.3} & 19.1         & 9.1  & \textbf{4.8} & 6.9          & 10.6 \\
\textcolor{CadetBlue}{Chain*} & 51.4         & \textbf{3.9} & ---   & 29.6         & \textbf{0.6} & ---   & 12.3         & \textbf{2.6} & ---   & 3.3          & \textbf{1.6} & ---   & 51.9         & \textbf{4.5} & ---   & \textbf{0.7} & 6.7          & ---   & 24.9         & \textbf{3.3} & ---   \\
\midrule
\textbf{Avg In-Domain}        & 4.6 & \textbf{0.4} & 14.8 & 6.6 & \textbf{4.0} & 14.6 & \textbf{2.7} & 3.4 & 15.9 & \textbf{1.7} & 3.9 & 16.2 & \textbf{3.8} & 5.3 & 13.1 & \textbf{1.9} & 7.9 & 17.2 & \textbf{3.5} & 4.2 & 15.3 \\
\bottomrule
\end{tabular}
\begin{tablenotes}
\small
\item * out-of-domain; $\Phi$-Random not run on chain for safety.
\end{tablenotes}
\end{threeparttable}}
\end{table}

\textbf{Secondary Tasks: Lobbing and Draping.}
Each multi-objective task uses two success metrics (Appendix Tab.~\ref{tab:combined_tasks}). \textbf{Lobbing}: a full success requires the rope tip to knock down a domino \textit{and} remain on the board (a gentle, controlled placement); the domino is positioned closer to the robot than the strike target to require overshoot. \textbf{Draping}: a success requires the tip to clear the wall and hang on the far side; secondary success uses the rope midpoint distance to the wall top, matching the CMA-ES-traj reward. Because $\Phi$ is shared across the task corpus, adding a new task requires only a representative simulation environment and a reward function, no retraining. Across the four in-domain ropes, $\Phi$-NN achieves a 54\% lobbing target-hit rate (21\% full success including the stay condition) and a 63\% draping success rate at 4.6\,cm mean midpoint distance; $\Phi$-CMA-ES is comparable (63\%/38\% lobbing, 67\%/4.1\,cm draping). The shared identification module thus supports qualitatively different downstream tasks (fast strikes, controlled placement, over-the-wall drapes) without any task-specific retraining.

\section{Limitations}
\label{sec:limitations}
 
CMA-ES-traj under Drake is CPU-bound, capping per-task throughput at 25 to 120 minutes depending on task and environment complexity (Sec.~\ref{sec:cma-es-traj-eval}); as discussed in Sec.~\ref{sec:cmaes-traj}, $\Phi$-NN's task-agnostic output is theoretically compatible with any future GPU-accelerated or gradient-based optimizer. An RL policy conditioned on $\hat{\xi}$ could amortize this per-task cost into a single training phase and run in milliseconds at deployment, and is an exciting future direction. $\Phi$-NN's predictions saturate at the bounds of its training distribution (Appendix~\ref{sec:ood_appendix}), which limits generalization to objects far outside the rope regime (like the chain).

Segmentation and tracking introduce noticeable noise in the real-to-sim transfer, and our current simulation has limited fidelity for behaviors like rope kinks.
 
\section{Conclusion}
 
This paper introduces a novel, task-agnostic system identification method for rope manipulation. Through the use of a single safe action and the leveraging of simulation priors with a neural network model we can fit descriptive system parameters to a rope in real. On 3D pole-striking this yields roughly a four-fold improvement in real-world accuracy over non-system-identified baselines, and the same identification module extends without modification to lobbing and draping. Targeted wiggles that excite individual parameters in isolation is a promising direction for future work.
\newpage

\acknowledgments{ This work used computational resources at Pittsburgh Supercomputing Center through allocation CIS251423 from the Advanced Cyberinfrastructure Coordination Ecosystem: Services \& Support (ACCESS) program, which is supported by U.S. National Science Foundation grants \#2138259, \#2138286, \#2138307, \#2137603, and \#2138296. \cite{accessgrant} 

We also thank Professor Chris Atkeson for his advice on the paper and Alan Wang for his early work on this project.}

\bibliography{references}  
\newpage
\appendix
\section*{Appendix}
\addcontentsline{toc}{section}{Appendix}

\section{Action Policy Optimization: Simulation Pipeline Results}
\label{sec:sim_full_appendix}

To benchmark CMA-ES-traj's performance with ground-truth, predicted, and random parameters, we evaluate the full pipeline on 100 simulated ropes (Table~\ref{tab:sim-full}). Trajectories optimized with ground-truth parameters reach 1.4\,cm mean error (1.2\,cm median); predicted parameters reach 5.7\,cm mean (2.1\,cm median); random in-distribution parameters reach 16.3\,cm mean (12.8\,cm median). This confirms that CMA-ES-traj reliably finds near-optimal trajectories given accurate parameters, isolating SysId quality as the dominant factor in real-world residual error.

\begin{table}[h]
\centering
\small
\caption{Average of 100 adaptation-predicted ropes on full pipeline in simulation. Comparison of ground truth, predicted and 100 random in-distribution ropes. Here, CMA-ES-traj was optimized until 25 iterations \textit{or} min-distance of $<$1.5cm.}
\label{tab:sim-full}
\begin{tabular}{
    l 
    S[table-format=2.1] 
    S[table-format=1.2] 
    S[table-format=3.1] 
    S[table-format=2.1]
    @{\,${}\pm{}$\,}
    S[table-format=2.1]
}
\toprule
 & {\textbf{Median}} & {\textbf{Min}} & {\textbf{Max}} & \multicolumn{2}{c}{\textbf{Mean}} \\
\midrule
\textbf{Ground Truth} & 1.2 & 0.25 & 21.6 & 1.4 & 2.2 \\
\textbf{Predicted} & 2.1 & 0.27 & 70.2 & 5.7 & 11.5 \\
\textbf{100 random ropes} & 12.8 & 0.14 & 104.3 & 16.3 & 12.1 \\
\bottomrule
\end{tabular}
\end{table}

\section{Parameter Importance Ablation Details}
\label{sec:param_importance_appendix}
This appendix expands the parameter-importance analysis referenced in Section~\ref{sec:param_importance}. Table~\ref{tab:param_importance} reports mean 3D tip-to-target error in simulation for five identification conditions: full $\Phi$-NN; three partial-identification settings where only a subset of $\Phi$-NN's predictions are kept and the rest are sampled at random from the training distribution (length and lead mass together, length only, lead mass only); and fully random parameters. Each row averages 10 random parameter draws per configuration over the 30 rope configurations used in our real-world evaluation. The full pipeline reaches 0.9\,cm mean error; supplying only the easily-measured length and lead-mass predictions closes only 3.3\,cm of the 13.3\,cm gap to fully random parameters, leaving 10.9\,cm of residual error driven by the remaining seven (stiffness, damping, mass-distribution, link-count) parameters that the wiggle identifies.

\begin{table}[h]
\centering
\small
\caption{Parameter importance ablation. Each row replaces a subset of $\Phi$-NN's nine predictions with random in-distribution values, keeping $\Phi$-NN's predictions only for the parameters listed. Evaluated in simulation over 10 random parameter sets per each of 30 rope configurations from our real-world evaluation.}
\label{tab:param_importance}
\begin{tabular}{l
    S[table-format=2.1]
    S[table-format=2.1]}
\toprule
\textbf{Condition (simulation)} & \textbf{Mean (\si{cm})} & \textbf{Std (\si{cm})} \\
\midrule
$\Phi$-NN (full pipeline)                 & 0.9  & 0.4  \\
Random + $\Phi$-NN length \& lead mass    & 10.9 & 10.1 \\
Random + $\Phi$-NN length only            & 13.3 & 13.5 \\
Random + $\Phi$-NN lead mass only         & 12.3 & 10.5 \\
Fully random in-distribution              & 14.2 & 12.5 \\
\bottomrule
\end{tabular}
\end{table}

\section{Secondary Task Results: Lobbing and Draping}
\label{sec:secondary_tasks_appendix}
This appendix expands the secondary-task results referenced in Section~\ref{sec:fullpipeline}. Table~\ref{tab:combined_tasks} reports per-rope success rates and distances on the lobbing and draping tasks, averaged over 6 configurations per rope (45/55/65\,cm lengths with 5\,g lead, plus 45\,cm with 10/20/30\,g leads). Across the four in-domain ropes, the in-domain averages are: lobbing target-hit 54\% ($\Phi$-NN) vs.\ 63\% ($\Phi$-CMA-ES); lobbing stay 21\% vs.\ 38\%; draping success 63\% vs.\ 67\%; and draping midpoint distance 4.6\,cm vs.\ 4.1\,cm. Both methods exceed 50\% success on draping for all four ropes and on lobbing target-hit for three of four---the Orange rope (heavy, stiff polyester) is the hardest, with both methods at 17\% lobbing target-hit and 50\% draping success. The complete per-configuration results, including which configurations achieved both lobbing sub-goals and the per-configuration draping midpoint distances, are in Table~\ref{tab:full_results}.

\begin{table}[h]
\footnotesize
\centering
\caption{Performance across lobbing and draping tasks. Each metric shows both methods side-by-side: \textbf{NN}~=~$\Phi$-NN, \textbf{CMA}~=~$\Phi$-CMA-ES. Best per (rope, metric) in \textbf{bold}. Arrows: \,$\uparrow$\,=\,higher better, \,$\downarrow$\,=\,lower better. Averaged over 6 configurations per rope; full per-configuration results in Tab.~\ref{tab:full_results}.}
\label{tab:combined_tasks}
\setlength{\tabcolsep}{6pt}
\resizebox{\textwidth}{!}{%
\begin{tabular}{l cc cc cc cc}
\toprule
 & \multicolumn{4}{c}{\textbf{Lobbing}} & \multicolumn{4}{c}{\textbf{Draping}} \\
\cmidrule(lr){2-5} \cmidrule(lr){6-9}
 & \multicolumn{2}{c}{Target $\uparrow$} & \multicolumn{2}{c}{Stay $\uparrow$} & \multicolumn{2}{c}{Success $\uparrow$} & \multicolumn{2}{c}{Distance (cm) $\downarrow$} \\
\cmidrule(lr){2-3} \cmidrule(lr){4-5} \cmidrule(lr){6-7} \cmidrule(lr){8-9}
\textbf{Rope} & NN & CMA & NN & CMA & NN & CMA & NN & CMA \\
\midrule
\textcolor{Brown}{Brown}      & 67\%          & \textbf{83\%} & \textbf{50\%} & 33\%          & 67\%          & \textbf{83\%} & 4.1          & \textbf{3.8} \\
\textcolor{Dandelion}{Yellow} & 67\%          & 67\%          & 17\%          & \textbf{50\%} & 67\%          & 67\%          & 7.4          & \textbf{4.1} \\
\textcolor{Red}{Red}          & 67\%          & \textbf{83\%} & 17\%          & \textbf{50\%} & 67\%          & 67\%          & \textbf{3.6} & 4.8 \\
\textcolor{Orange}{Orange}    & 17\%          & 17\%          & 0\%           & \textbf{17\%} & 50\%          & 50\%          & \textbf{3.4} & 3.5 \\
\midrule
\textbf{Avg In-Domain}        & 54\%          & \textbf{63\%} & 21\%          & \textbf{38\%} & 63\%          & \textbf{67\%} & 4.6          & \textbf{4.1} \\
\bottomrule
\end{tabular}}
\end{table}

\section{Real-World Deployment Pipeline}
\label{sec:deployment_appendix}

Our complete system is designed for real-world deployment on physical robots for ropes with unknown quantitative parameters. The deployment pipeline is as follows: (1) execute wiggle and extract visual observations, (2) have $\Phi$-NN estimate rope parameters $\hat{\xi}$, (3) optimize trajectory using $\hat{\xi}$ in simulation, (4) execute trajectory on the robot, and (5) measure task performance. The key challenges are obtaining visual observations compatible with $\Phi$'s training data and accurately measuring task error in metric space.

\textbf{Wiggle Observation.} A ZED Mini 2i camera is positioned to match the simulation's pinhole calibration. We segment the first frame with Grounding SAM~\cite{ren2024grounded} and color refinement, fit a spline to the rope mask, and resample keypoints along the centerline at uniform arc-length. CoTracker~\cite{karaev2410cotracker3} tracks these keypoints across the wiggle, after which we convert to the same normalized angular features used in training. For real-world deployment, we use only position and angle features ($\tilde{p}_t$, $\theta_t$) to avoid noise amplification from temporal derivatives. Domain randomization during $\Phi$'s training (calibration noise, tracking noise) ensures robustness to real-world imperfections.

\textbf{Trajectory Execution and Evaluation.} After $\Phi$-NN predicts parameters ($\hat{\xi}$), we optimize a trajectory in Drake with CMA-ES as explained in Sec.~\ref{sec:cmaes-traj} and execute it on the xArm 7. To measure task error, we track the rope endpoint and target with HSV-based color filtering. The ZED camera's depth estimates convert 2D detections to 3D metric coordinates, yielding the minimum distance between rope endpoint and target.

\section{Feature Engineering Details}
\label{sec:features_appendix}

We normalize positions relative to the first link: $\tilde{p}_t^{(i)} = p_t^{(i)} - p_t^{(1)}$. This provides two benefits: (1) the first link is attached to the robot end-effector (effectively a fixed hinge), conveying minimal rope-specific information, and (2) normalization provides robustness to calibration shifts between simulation and deployment. We then compute angular features from normalized positions, including angles relative to the first link and their temporal derivatives (angular velocity and acceleration) using centered finite differences with Gaussian smoothing. While derivative features improve performance in simulation, they amplify noise in real-world due to tracking errors. Therefore, for simulation training we use only position and angle features: $\mathcal{O} = \{\tilde{p}_t, \theta_t\}$ to ensure minimal feature drift during real world deployment.

\begin{table}[h]
\centering
\small
\caption{Complete experimental results across all rope configurations and parameter prediction methods. Results were repeatable, so one trial was performed for each task-rope combination.}
\label{tab:full_results}
\resizebox{\textwidth}{!}{%
\begin{threeparttable}
\begin{tabular}{@{}ll@{\hspace{3pt}}c@{}c@{\hspace{4pt}}c@{}c@{\hspace{4pt}}c@{}c@{\hspace{4pt}}c@{}c@{\hspace{4pt}}c@{}c@{\hspace{4pt}}c@{}c@{\hspace{4pt}}@{\hspace{4pt}}c@{}c@{\hspace{4pt}}c@{}c@{\hspace{4pt}}c@{}c@{\hspace{4pt}}c@{}c@{\hspace{4pt}}c@{}c@{\hspace{4pt}}c@{}c@{}}
\toprule
& & \multicolumn{12}{c@{\hspace{10pt}}}{\textbf{Lobbing}} & \multicolumn{12}{c}{\textbf{Draping}} \\
\cmidrule(lr){3-14} \cmidrule(l){15-26}
& & \multicolumn{2}{c@{\hspace{4pt}}}{45cm} & \multicolumn{2}{c@{\hspace{4pt}}}{45cm} & \multicolumn{2}{c@{\hspace{4pt}}}{45cm} & \multicolumn{2}{c@{\hspace{4pt}}}{45cm} & \multicolumn{2}{c@{\hspace{4pt}}}{55cm} & \multicolumn{2}{c@{\hspace{4pt}}}{65cm} 
& \multicolumn{2}{c@{\hspace{4pt}}}{45cm} & \multicolumn{2}{c@{\hspace{4pt}}}{45cm} & \multicolumn{2}{c@{\hspace{4pt}}}{45cm} & \multicolumn{2}{c@{\hspace{4pt}}}{45cm} & \multicolumn{2}{c@{\hspace{4pt}}}{55cm} & \multicolumn{2}{c}{65cm} \\
\textbf{Rope} & \textbf{Method} & \multicolumn{2}{c@{\hspace{4pt}}}{5g} & \multicolumn{2}{c@{\hspace{4pt}}}{10g} & \multicolumn{2}{c@{\hspace{4pt}}}{20g} & \multicolumn{2}{c@{\hspace{4pt}}}{30g} & \multicolumn{2}{c@{\hspace{4pt}}}{5g} & \multicolumn{2}{c@{\hspace{10pt}}}{5g}
& \multicolumn{2}{c@{\hspace{4pt}}}{5g} & \multicolumn{2}{c@{\hspace{4pt}}}{10g} & \multicolumn{2}{c@{\hspace{4pt}}}{20} & \multicolumn{2}{c@{\hspace{4pt}}}{30g} & \multicolumn{2}{c@{\hspace{4pt}}}{5g} & \multicolumn{2}{c}{5g} \\
\cmidrule(lr){3-4} \cmidrule(lr){5-6} \cmidrule(lr){7-8} \cmidrule(lr){9-10} \cmidrule(lr){11-12} \cmidrule(lr){13-14}
\cmidrule(lr){15-16} \cmidrule(lr){17-18} \cmidrule(lr){19-20} \cmidrule(lr){21-22} \cmidrule(lr){23-24} \cmidrule(l){25-26}
& & \textbf{T} & \textbf{S} & \textbf{T} & \textbf{S} & \textbf{T} & \textbf{S} & \textbf{T} & \textbf{S} & \textbf{T} & \textbf{S} & \textbf{T} & \textbf{S}
& \textbf{S} & \textbf{D} & \textbf{S} & \textbf{D} & \textbf{S} & \textbf{D} & \textbf{S} & \textbf{D} & \textbf{S} & \textbf{D} & \textbf{S} & \textbf{D} \\
\midrule
\multirow{2}{*}{\textcolor{Brown}{Brown}} 
 & $\Phi$-NN & \cmark & --- & \cmark & \cmark & \cmark & \cmark & --- & --- & --- & --- & \cmark & \cmark 
 & \cmark & 3.5 & \cmark & 5.0 & \cmark & 5.0 & \cmark & 3.0 & --- & --- & --- & --- \\
 & $\Phi$-CMA-ES & \cmark & --- & \cmark & --- & --- & --- & \cmark & --- & \cmark & \cmark & \cmark & \cmark 
 & \cmark & 4.5 & \cmark & 4.5 & \cmark & 4.5 & \cmark & 5.5 & \cmark & 0.0 & --- & --- \\
 \addlinespace
\multirow{2}{*}{\textcolor{Dandelion}{Yellow}} 
 & $\Phi$-NN & --- & --- & \cmark & --- & \cmark & --- & \cmark & --- & --- & --- & \cmark & \cmark 
 & \cmark & 10.0 & \cmark & 8.5 & \cmark & 11.0 & --- & 11.0 & \cmark & 0.0 & --- & --- \\
 & $\Phi$-CMA-ES & \cmark & \cmark & --- & --- & --- & --- & \cmark & --- & \cmark & \cmark & \cmark & \cmark 
 & \cmark & 3.5 & \cmark & 4.0 & \cmark & 3.5 & --- & 6.0 & \cmark & 5.5 & --- & --- \\
 \addlinespace
\multirow{2}{*}{\textcolor{Red}{Red}} 
 & $\Phi$-NN & --- & --- & \cmark & --- & \cmark & --- & \cmark & --- & \cmark & \cmark & --- & --- 
 & \cmark & 4.5 & \cmark & 4.5 & \cmark & 3.5 & \cmark & 1.8 & --- & --- & --- & --- \\
 & $\Phi$-CMA-ES & \cmark & \cmark & \cmark & --- & --- & --- & \cmark & --- & \cmark & \cmark & \cmark & \cmark 
 & \cmark & 9.5 & \cmark & 4.5 & \cmark & 3.0 & \cmark & 2.3 & --- & --- & --- & --- \\
 \addlinespace
\multirow{2}{*}{\textcolor{Orange}{Orange}} 
 & $\Phi$-NN & --- & --- & --- & --- & --- & --- & --- & --- & --- & --- & \cmark & --- 
 & \cmark & 3.9 & \cmark & 4.8 & --- & --- & --- & --- & \cmark & 1.5 & --- & --- \\
 & $\Phi$-CMA-ES & --- & --- & --- & --- & --- & --- & --- & --- & --- & --- & \cmark & \cmark 
 & \cmark & 4.5 & \cmark & 5.4 & --- & --- & --- & --- & \cmark & 0.5 & --- & --- \\
\bottomrule
\end{tabular}
\begin{tablenotes}
\small
\item T = Target hit, S = Stayed, D = Distance (cm). \cmark~= success, ---~= failure or not measured.
\end{tablenotes}
\end{threeparttable}
}
\end{table}

\section{Wiggle Ablation Study}
\label{sec:wiggleablation_appendix}

We validate that the specific wiggle trajectory used for system identification is not critical by training 8 additional models with different planar wiggles: 6 systematically varying amplitude (20$^\circ$, 30$^\circ$) and frequency (0.5--1.0\,Hz), and 2 random predefined trajectories (Table~\ref{tab:wiggle_ablation}). The main wiggle achieves best overall performance, with Ablation 1 (20$^\circ$, 0.5\,Hz) close behind. Performance remains consistent across planar wiggles that sufficiently excite rope dynamics, and no single wiggle dominates all parameters. Random trajectories (Ablations 7--8) show degraded performance on complex parameters. This validates our design principle: planar wiggles with adequate excitation enable accurate parameter inference regardless of specific trajectory. We hypothesize that stereo setups with depth cameras and non-planar motion could further improve inference of certain parameters like damping and stiffness. Per-parameter gradient sensitivity maps (Appendix Tab.~\ref{tab:activation_heatmaps}) further suggest that different parameters are informed by different timesteps of the wiggle: inertial and damping parameters (ball stiffness, ball damping, mass per unit length) peak at later trajectory inflection points where dynamically similar ropes diverge, while geometric parameters are identifiable earlier. Sensitivity diminishes toward the sequence end, indicating that wiggle duration is sufficient but not over-utilized.
\linebreak
\begin{table*}[h]
\centering
\footnotesize
\caption{Wiggle Study: Mean Absolute Error across wiggle variants (1000 test samples). Ablations systematically vary wiggle parameters: Abl 1-3 use 20$^\circ$ amplitude at 0.5, 0.75, 1.0 Hz; Abl 4-6 use 30$^\circ$ amplitude at 0.5, 0.75, 1.0 Hz; Abl 7-8 use random predefined trajectories. Main is the primary wiggle (joint 6 oscillation of the xArm) used throughout all experiments. Color gradient indicates relative performance per parameter (green = best, red = worst); \textbf{bold} marks the best wiggle for each parameter (ties bolded).}
\label{tab:wiggle_ablation}
\resizebox{\textwidth}{!}{
\begin{tabular}{l|c|c|c|c|c|c|c|c|c}
\toprule
\textbf{Parameter} & \textbf{Abl 1} & \textbf{Abl 2} & \textbf{Abl 3} & \textbf{Abl 4} & \textbf{Abl 5} & \textbf{Abl 6} & \textbf{Abl 7} & \textbf{Abl 8} & \textbf{Main} \\
 & \scriptsize{20$^\circ$, 0.5Hz} & \scriptsize{20$^\circ$, 0.75Hz} & \scriptsize{20$^\circ$, 1.0Hz} & \scriptsize{30$^\circ$, 0.5Hz} & \scriptsize{30$^\circ$, 0.75Hz} & \scriptsize{30$^\circ$, 1.0Hz} & \scriptsize{Random 1} & \scriptsize{Random 2} & \scriptsize{Joint 6} \\
\midrule
Number of links & \cellcolor{green!40}0.122 & \cellcolor{green!30}0.127 & \cellcolor{yellow!30}0.225 & \cellcolor{yellow!20}0.205 & \cellcolor{yellow!20}0.206 & \cellcolor{orange!30}0.360 & \cellcolor{red!50}2.098 & \cellcolor{red!30}1.453 & \cellcolor{green!60}\textbf{0.098} \\
Rope length (m) & \cellcolor{green!40}0.007 & \cellcolor{green!20}0.013 & \cellcolor{yellow!30}0.017 & \cellcolor{yellow!20}0.015 & \cellcolor{yellow!30}0.018 & \cellcolor{orange!20}0.022 & \cellcolor{red!50}0.057 & \cellcolor{red!30}0.028 & \cellcolor{green!60}\textbf{0.006} \\
Ball damping (N$\cdot$s/m) & \cellcolor{green!60}\textbf{0.010} & \cellcolor{green!20}0.033 & \cellcolor{yellow!30}0.038 & \cellcolor{green!30}0.018 & \cellcolor{green!20}0.031 & \cellcolor{red!50}0.096 & \cellcolor{yellow!30}0.036 & \cellcolor{orange!30}0.040 & \cellcolor{green!60}\textbf{0.010} \\
Rope radius (m) & \cellcolor{green!60}\textbf{0.002} & \cellcolor{green!20}0.003 & \cellcolor{green!20}0.003 & \cellcolor{green!20}0.003 & \cellcolor{yellow!30}0.004 & \cellcolor{red!50}0.006 & \cellcolor{green!20}0.003 & \cellcolor{orange!30}0.004 & \cellcolor{green!60}\textbf{0.002} \\
Mass per unit length (kg/m) & \cellcolor{green!60}\textbf{0.006} & \cellcolor{green!20}0.010 & \cellcolor{green!20}0.010 & \cellcolor{yellow!30}0.013 & \cellcolor{yellow!20}0.012 & \cellcolor{orange!20}0.014 & \cellcolor{red!50}0.019 & \cellcolor{orange!30}0.014 & \cellcolor{green!40}0.007 \\
Link extra scale & \cellcolor{green!40}0.153 & \cellcolor{yellow!20}0.160 & \cellcolor{yellow!20}0.160 & \cellcolor{yellow!30}0.179 & \cellcolor{yellow!30}0.169 & \cellcolor{red!50}0.241 & \cellcolor{orange!30}0.205 & \cellcolor{red!30}0.208 & \cellcolor{green!60}\textbf{0.152} \\
Lead mass (kg) & \cellcolor{green!60}\textbf{0.005} & \cellcolor{yellow!20}0.011 & \cellcolor{yellow!30}0.014 & \cellcolor{green!30}0.008 & \cellcolor{yellow!20}0.011 & \cellcolor{orange!30}0.018 & \cellcolor{red!30}0.021 & \cellcolor{red!50}0.016 & \cellcolor{green!60}\textbf{0.005} \\
Lead radius (m) & \cellcolor{green!20}0.008 & \cellcolor{green!20}0.008 & \cellcolor{green!20}0.008 & \cellcolor{green!20}0.008 & \cellcolor{green!20}0.008 & \cellcolor{green!20}0.008 & \cellcolor{green!60}\textbf{0.007} & \cellcolor{green!20}0.008 & \cellcolor{green!20}0.008 \\
Ball stiffness (N/m) & \cellcolor{green!40}0.118 & \cellcolor{orange!30}0.257 & \cellcolor{red!30}0.258 & \cellcolor{yellow!30}0.224 & \cellcolor{yellow!30}0.244 & \cellcolor{orange!30}0.336 & \cellcolor{red!50}0.388 & \cellcolor{red!30}0.322 & \cellcolor{green!60}\textbf{0.111} \\
\bottomrule
\end{tabular}}
\end{table*}

\section{Out-of-Distribution Robustness}
\label{sec:ood_appendix}

We test $\Phi$-NN's generalization by extending four parameters (ball damping, mass per unit length, lead mass, ball stiffness) beyond training bounds. The model clamps predictions to training-distribution bounds rather than extrapolating: ball damping predictions saturate at $[0.016, 0.089]$ when targets span $[0.100, 1.499]$. This is a fundamental limit of supervised models trained on bounded distributions.

\section{Transferability}~\label{sec:tranferability-appendix}
Appendix Table~\ref{tab:phi_comparison} provides the per-rope motion-fidelity results referenced in Section~\ref{sec:fullpipeline} of the main paper. We execute a higher-jerk motion in real and compare the rope's behavior to simulations parameterized by $\Phi$-NN, $\Phi$-CMA-ES, and a random in-distribution baseline. Across 10 in-distribution rope/lead pairs, $\Phi$-NN and $\Phi$-CMA-ES both achieve 0.95 average Fourier frequency correlation, versus 0.81 for random parameters. Per-frame point distance follows the same pattern: 5.4\,cm/frame ($\Phi$-NN), 5.8\,cm/frame ($\Phi$-CMA-ES), and 13.6\,cm/frame (random). The chain (out-of-distribution) is an exception: $\Phi$-CMA-ES's per-rope overfitting outperforms $\Phi$-NN's learned priors there. 

\begin{table}[h]
\centering
\caption{Estimation of parameters, material, and physical description of test ropes on different $\Phi$ methods.}
\label{tab:phi_comparison}
\begin{threeparttable}
\setlength{\tabcolsep}{4pt}
\begin{tabular}{llcccccc}
\toprule
\textbf{Rope} & \textbf{Lead} & \multicolumn{3}{c}{\textbf{Frequency correlation}} & \multicolumn{3}{c}{\textbf{Average Point Distance}**} \\
& \textbf{(g)} & NN & CMAES & Random* & NN & CMAES & Random* \\
\midrule
\multirow{2}{*}{\textcolor{Brown}{Brown}}  & 10 & 0.98 & 0.94 & 0.76 & 4.8 & 6.9 & 13.4 \\
 & 30 & 0.97 & 0.91 & 0.85 & 4.7 & 8.4 & 14.7 \\
\cmidrule(lr){1-8}
\multirow{2}{*}{\textcolor{Dandelion}{Yellow}} & 10 & 0.98 & 0.96 & 0.80 & 4.0 & 5.6 & 13.7 \\
 & 30 & 0.97 & 0.94 & 0.77 & 4.9 & 7.4 & 15.2 \\
\cmidrule(lr){1-8}
\multirow{2}{*}{\textcolor{Red}{Red}} & 10 & 0.97 & 0.95 & 0.81 & 7.7 & 5.8 & 14.2 \\
 & 30 & 0.97 & 0.93 & 0.78 & 6.3 & 8.3 & 15.9 \\
\cmidrule(lr){1-8}
\multirow{2}{*}{\textcolor{Orange}{Orange}} & 10 & 0.97 & 0.89 & 0.91 & 4.5 & 4.3 & 12.4 \\
 & 30 & 0.99 & 0.99 & 0.90 & 3.2 & 2.8 & 10.7 \\
\cmidrule(lr){1-8}
\multirow{2}{*}{\textcolor{CadetBlue}{Chain***}} & 10 & 0.77 & 0.99 & 0.76 & 8.4 & 3.2 & 14.2 \\
 & 30 & 0.98 & 0.99 & 0.78 & 6.0 & 5.2 & 12.8 \\
\midrule
Average &  & 0.95 & 0.95 & 0.81 & 5.4 & 5.8 & 13.6 \\
\cmidrule(lr){3-8}
Std.\ Dev. & & 0.06 & 0.03 & 0.05 & 1.7 & 2.0 & 1.6 \\
\bottomrule
\end{tabular}
\begin{tablenotes}
\small
\item * For $\Phi$-Random we randomly sample 10 (simulation stable) ropes as a control. Distance is mean trajectory error in the image plane (ZED Mini camera model, depth 0.65\,m), in centimeters.
\item ** cm per frame
\item *** out-of-distribution
\end{tablenotes}
\end{threeparttable}
\end{table}

\section{$\Phi$-NN Activations Analysis}

We perform a case study of $\Phi$-NN activations by computing gradient-based sensitivity $\frac{\partial p_i}{\partial \mathbf{x}_t}$ for each predicted parameter $p_i$ with respect to input joint positions $\mathbf{x}_t$ and angles $\theta_t$ at each timestep, revealing which spatiotemporal regions of the wiggle most influence each parameter's prediction (Appendix Table~\ref{tab:activation_heatmaps}). Ball stiffness, ball damping, and mass parameters exhibit their highest sensitivity later in the wiggle, with peaks at apparent trajectory inflection points where physical differences emerge most clearly. Sensitivity diminishes toward the sequence end, suggesting that while extended duration is needed to capture temporal dynamics, yet additional wiggle time may provide limited new information.
We hypothesize that multi-parameter relationships, such as stiff, light-lead ropes initially wiggling similarly to non-stiff, heavy-lead ropes before the latter gain momentum, contribute to these later activations.

\section{Implementation Details}\label{sec:imp_details}

\subsection{Network Architecture}
$\Phi$-NN uses a temporal convolutional encoder followed by a multi-layer perceptron. The encoder consists of three 1D convolutional blocks applied along the temporal dimension. Each block uses kernel size 8, stride 1, followed by layer normalization, GELU activation, average pooling (stride 2), and dropout (rate 0.3). After the convolutional blocks, adaptive pooling produces a fixed-length representation, which is passed through a linear layer to produce a 256-dimensional embedding. The MLP head uses hidden dimensions [128, 64] to map the embedding to 9 normalized rope parameters $\hat{\xi} \in [0,1]^9$. The complete model has 1.2M parameters.

\subsection{Feature Engineering}
Angular velocity is computed using unwrapped finite differences with Gaussian smoothing ($\sigma=1.0$, kernel size 5). Angular acceleration uses the same approach with $\sigma=1.5$ and kernel size 7. We record rope motion for 400 frames at 60 FPS ($\sim$6.7 seconds).

\subsection{Domain Randomization}
Calibration noise: Gaussian noise ($\sigma=2$\,cm) applied to camera position and lookat point. Tracking noise: Anisotropic noise spanning 0-3 pixels with temporal correlation coefficient $\alpha=0.8$, with higher variance in the longitudinal direction than lateral to mimic realistic tracking errors. Trajectory padding: Randomly add 0-20 frames at the start of each trajectory to simulate recording delays.

\subsection{Curriculum Masking Schedule}
Masking begins at epoch 50 using 50-frame contiguous blocks. From epochs 50-200, we mask 1-2 random blocks per trajectory. From epochs 200-400, we apply beginning-biased masking where blocks near the trajectory start are preferentially masked. From epochs 400-500, we increase to 7 masked blocks per trajectory. All masked frames are set to zero.

\subsection{Training Configuration}
Dataset: 9000 training ropes and 1000 validation ropes with parameters sampled using Latin Hypercube Sampling (LHS) from the ranges in Table II (Main Paper). Loss: Mean squared error on normalized parameters. Optimizer: Adam with initial learning rate $10^{-3}$, cosine annealing schedule, and 5-epoch warmup. Batch size: 32. Training duration: 500 epochs.

\subsection{Trajectory Parameterization and Velocity Filtering}
Trajectories are parameterized by three sequential joint-space waypoints, interpolated by a cubic spline to produce smooth continuous joint trajectories. For the full 7-DOF arm this gives 21 parameters; for the planar lobbing and draping tasks we restrict to the three in-plane joints (9 parameters). At sample time, each candidate's spline-implied joint velocity profile is evaluated against the xArm 7 joint-velocity limits, and samples that violate any joint's limit are rejected before fitness evaluation, so all evaluated trajectories are hardware-feasible. CMA-ES-traj runs up to 25 iterations of 60 samples each for pole striking, 50 iterations for lobbing, and 15 for draping. For pole striking the initial sampling distribution is unbiased; for lobbing and draping it is biased toward a lifting motion as a warm start.

\begin{table*}[t]
\centering
\caption{Neural network activation heatmaps showing gradient flow for each rope parameter during wiggle observation. These are for one demonstration of our 45cm brown rope with five gram lead.}
\label{tab:activation_heatmaps}
\begin{tabular}{@{}c@{\hspace{1.5em}}c@{\hspace{1.5em}}c@{}}
\toprule
Number of links & Rope radius & Lead mass \\
\midrule
\includegraphics[width=0.29\linewidth]{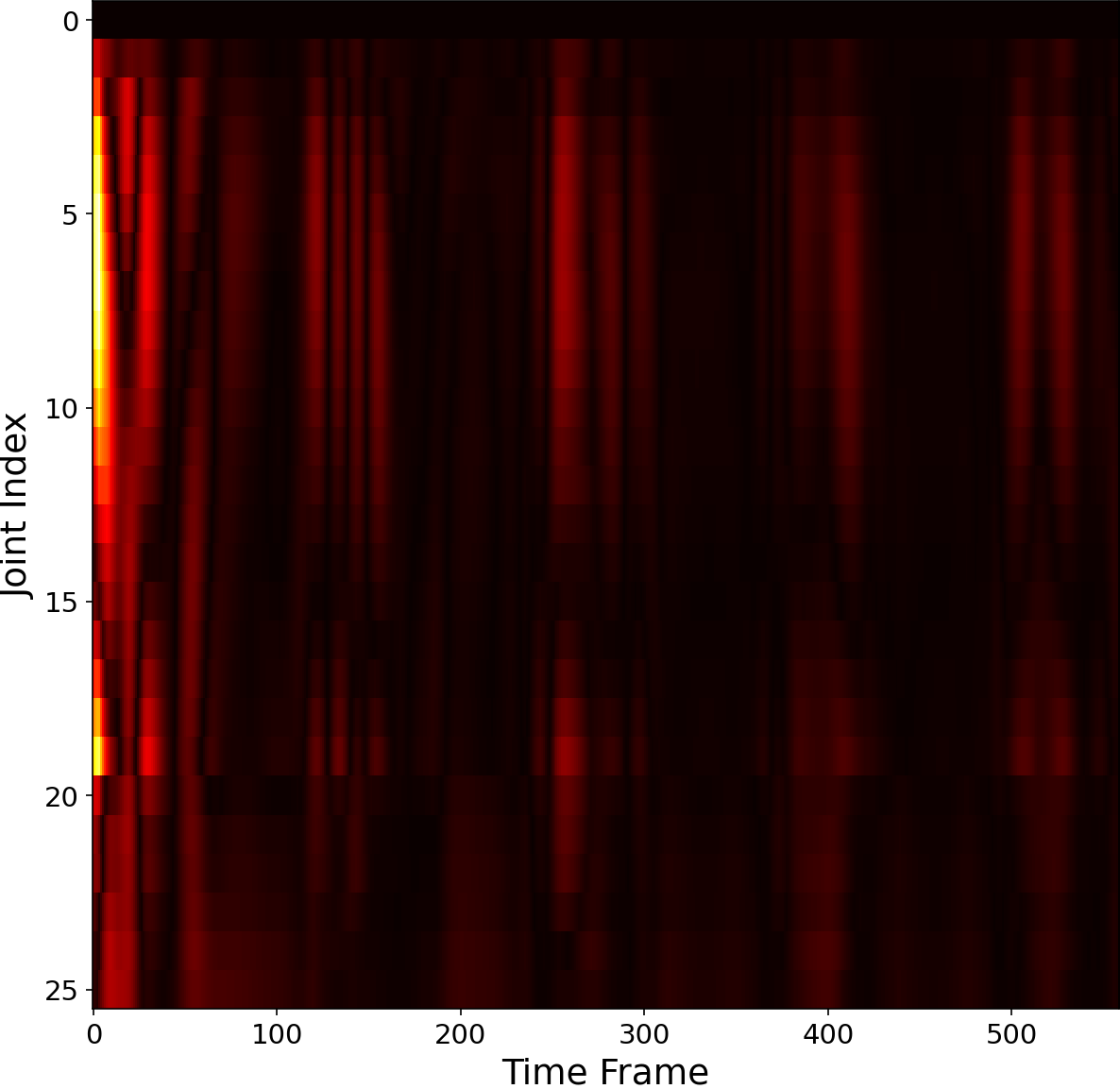} & 
\includegraphics[width=0.29\linewidth]{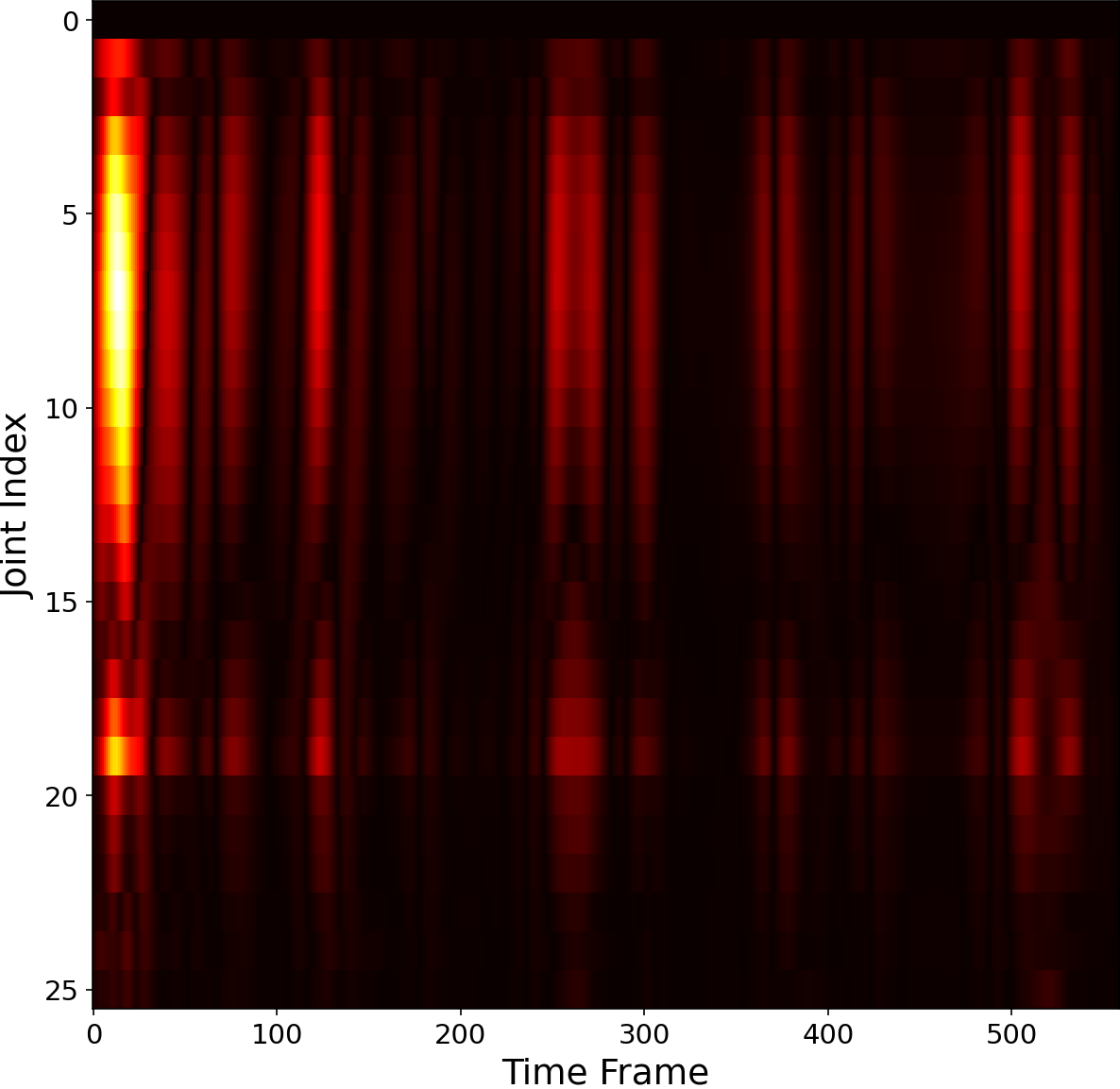} & 
\includegraphics[width=0.29\linewidth]{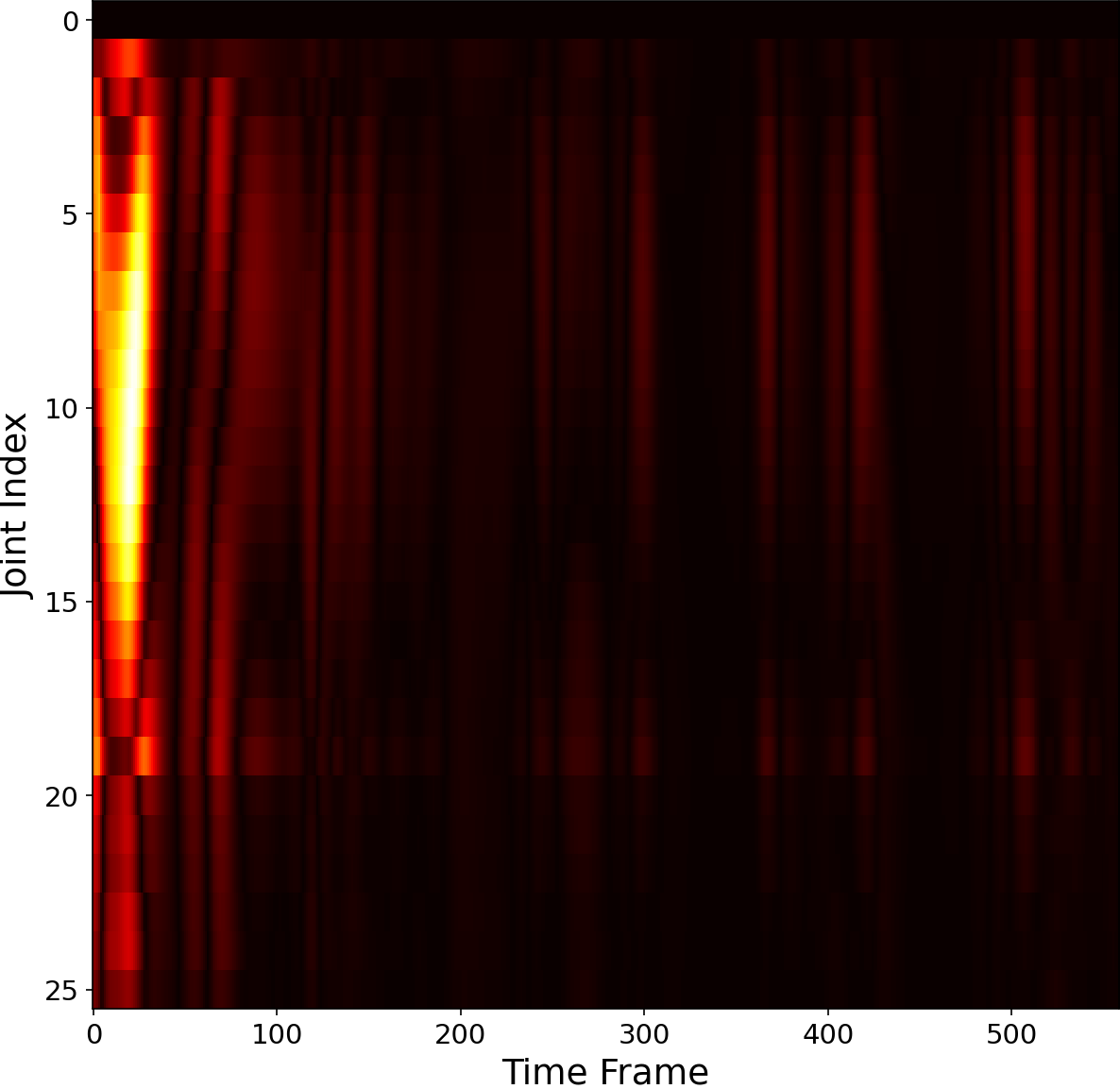} \\[1em]
\midrule
Rope length & Mass per unit len & Lead radius \\
\midrule
\includegraphics[width=0.29\linewidth]{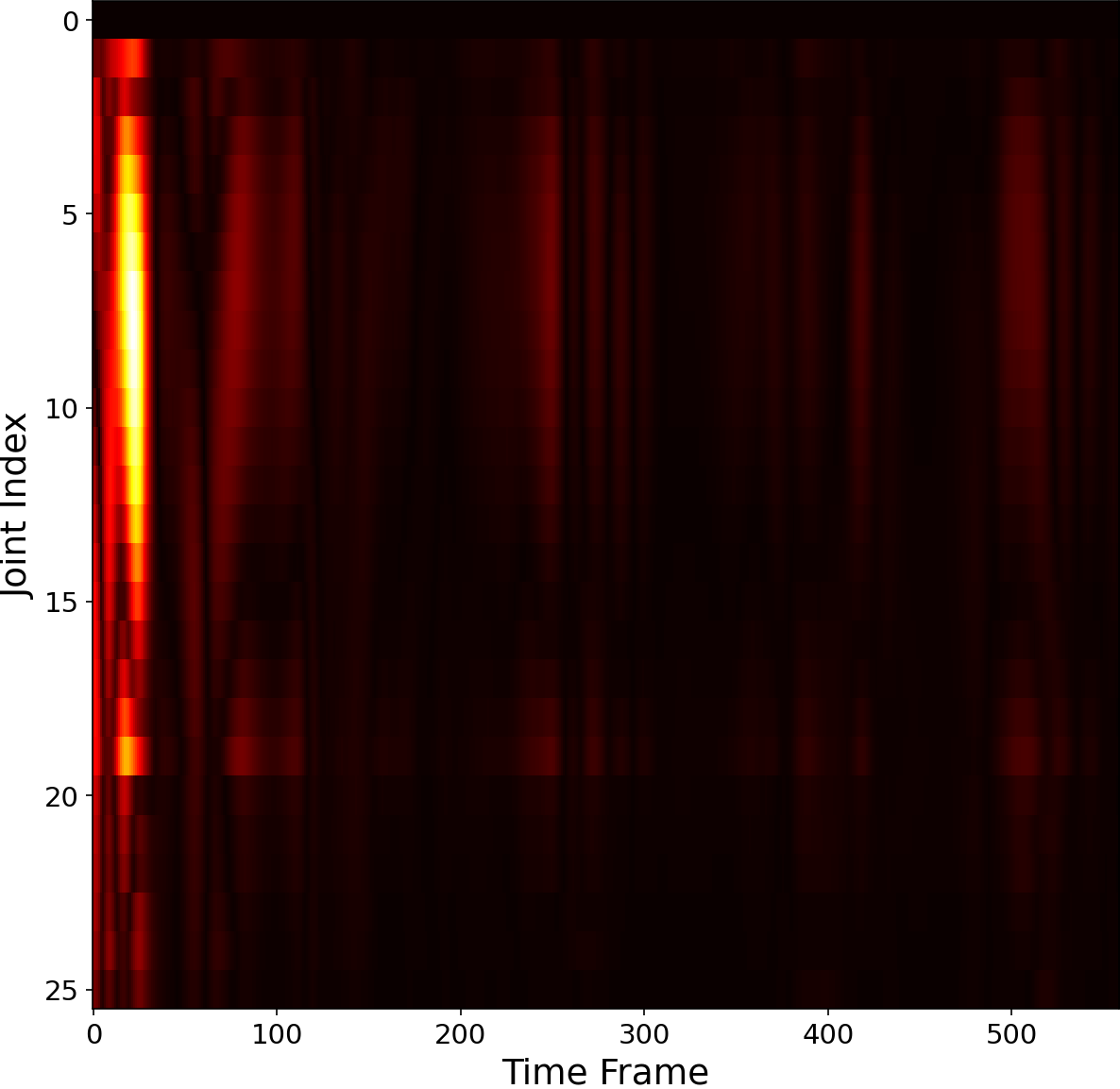} & 
\includegraphics[width=0.29\linewidth]{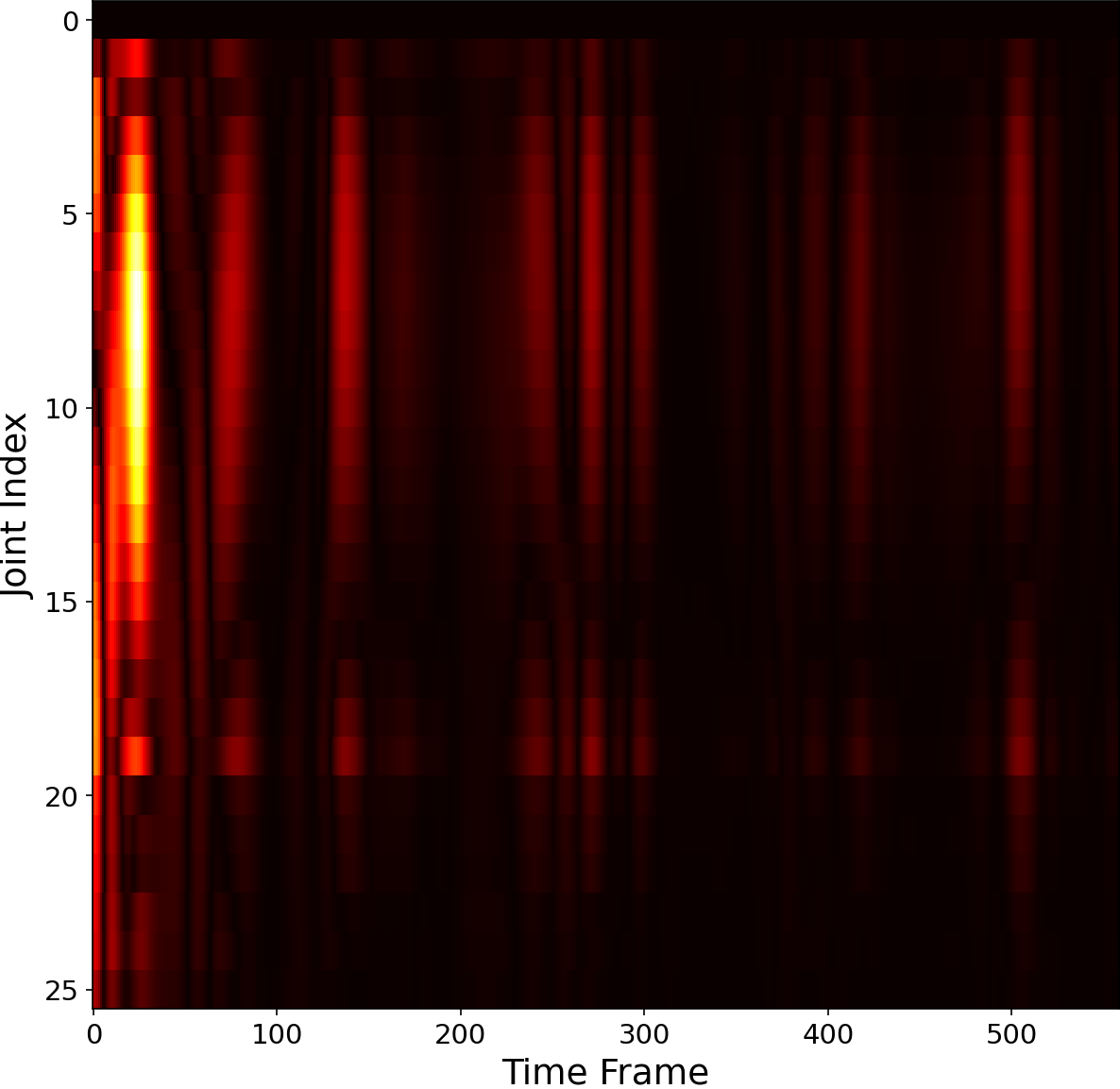} & 
\includegraphics[width=0.29\linewidth]{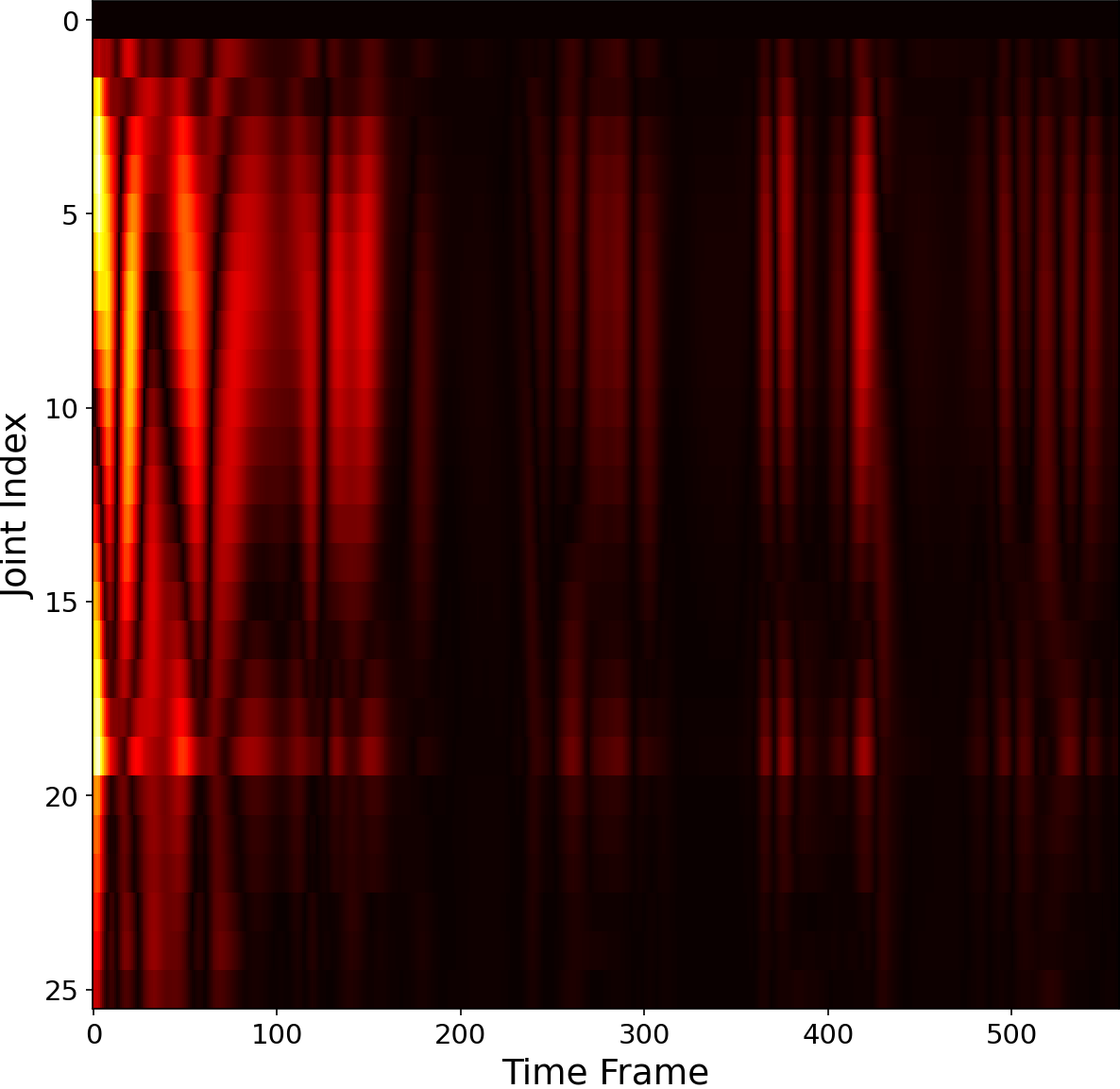} \\[1em]
\midrule
Ball damping & Ball stiffness & Link extra scale \\
\midrule
\includegraphics[width=0.29\linewidth]{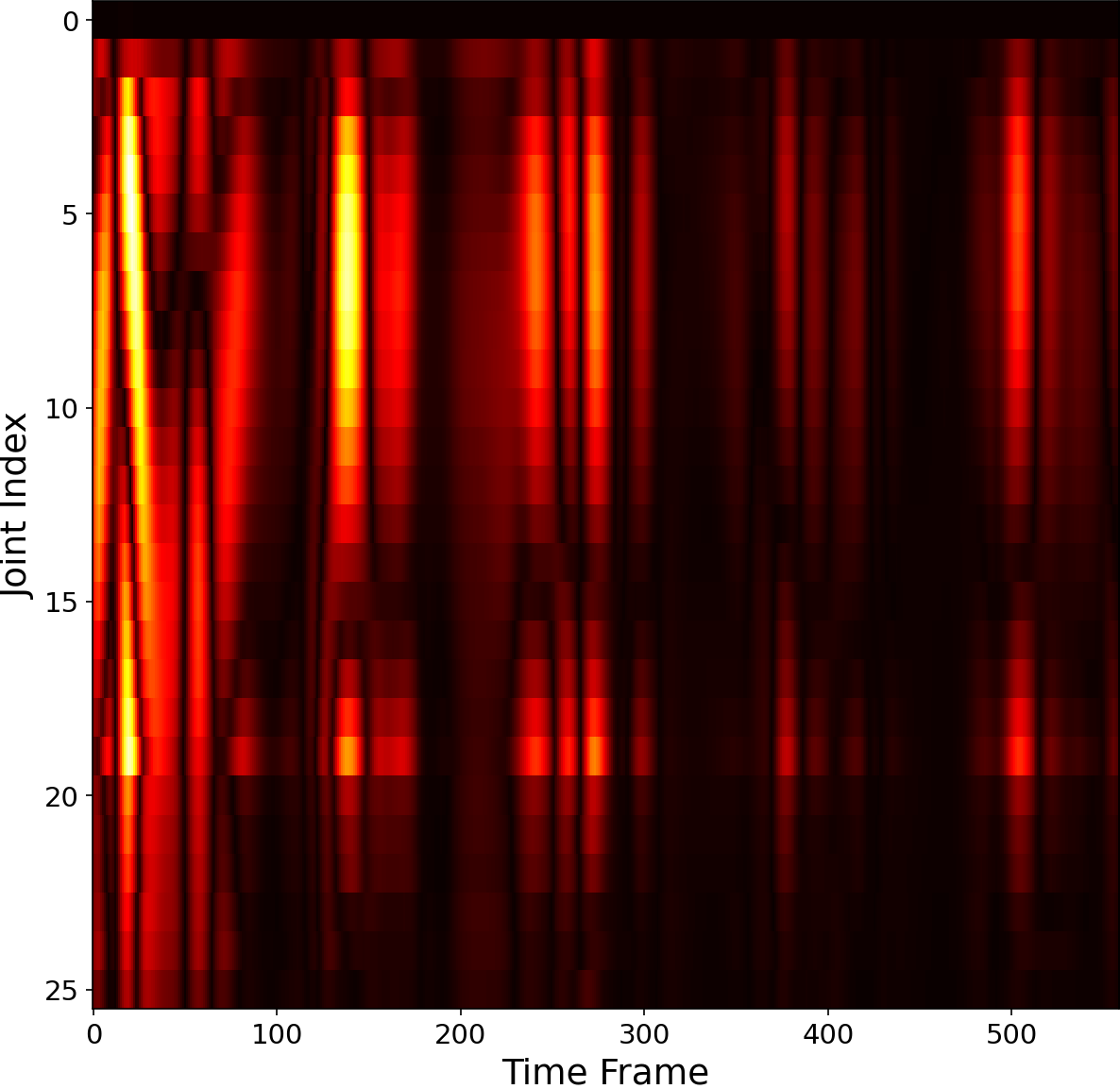} & 
\includegraphics[width=0.29\linewidth]{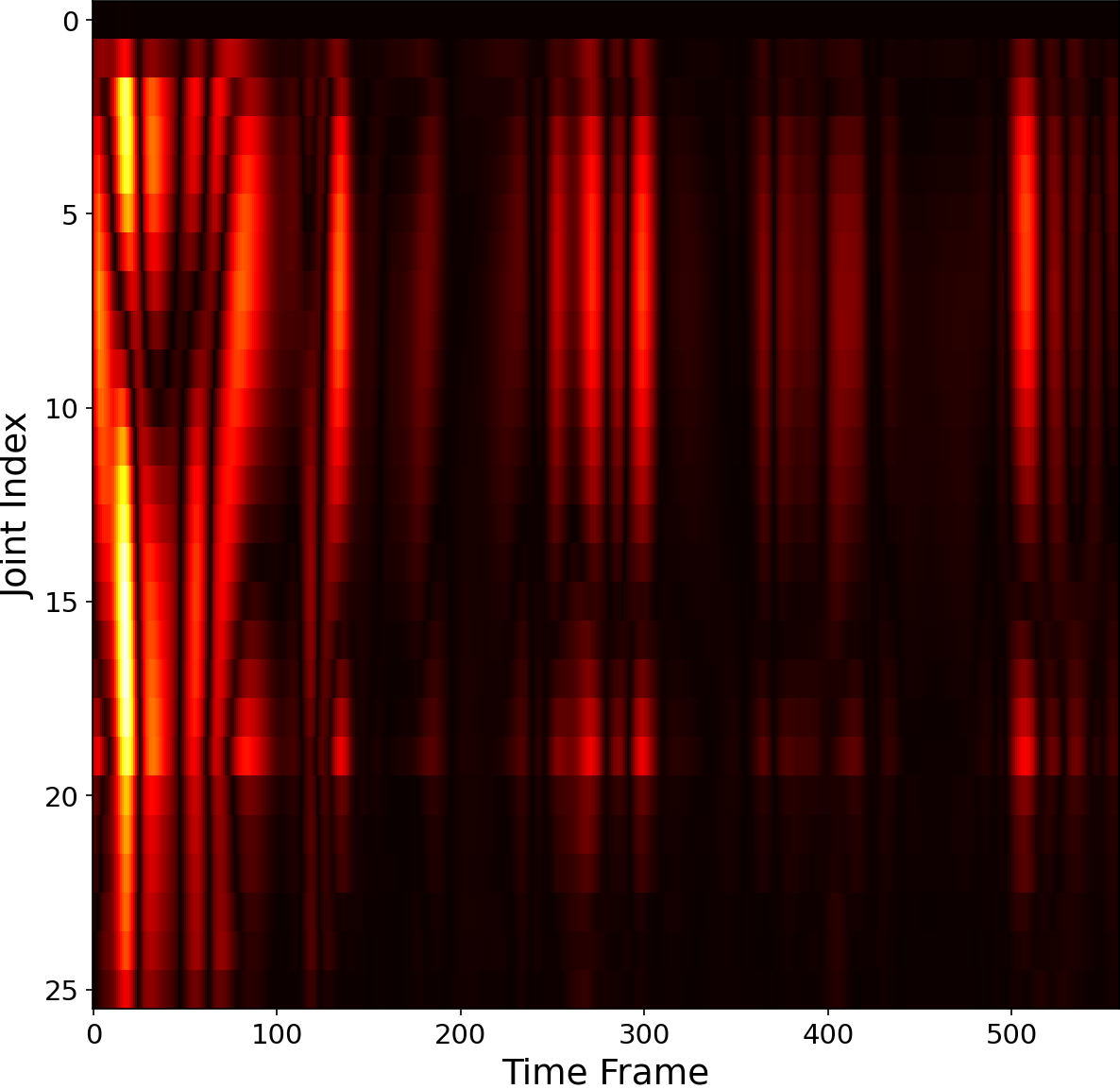} & 
\includegraphics[width=0.29\linewidth]{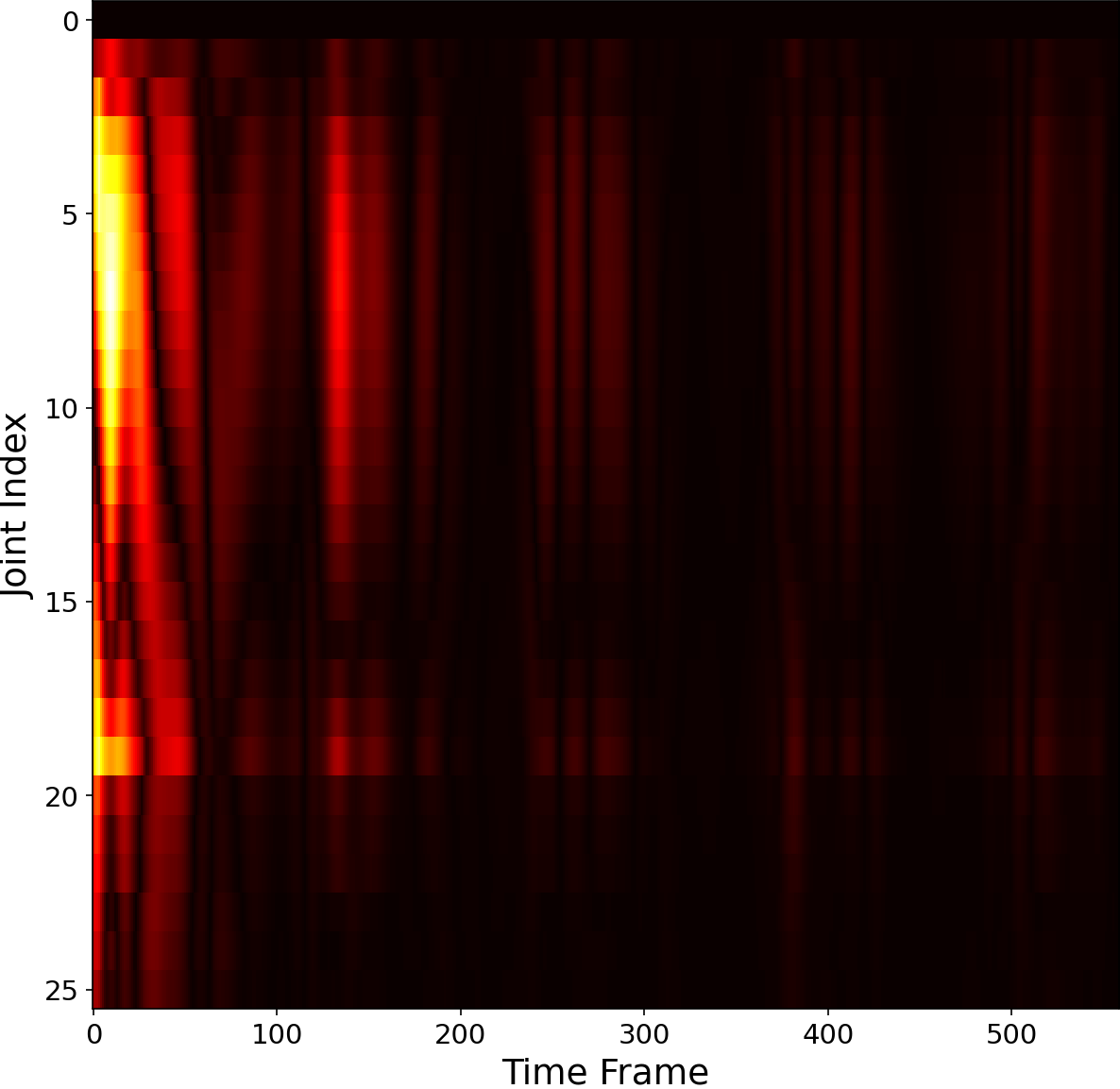} \\[0.5em]
\cmidrule{2-2}
\multicolumn{3}{c}{\textbf{All Parameters}} \\
\cmidrule{2-2}
\multicolumn{3}{c}{\includegraphics[width=0.30\linewidth]{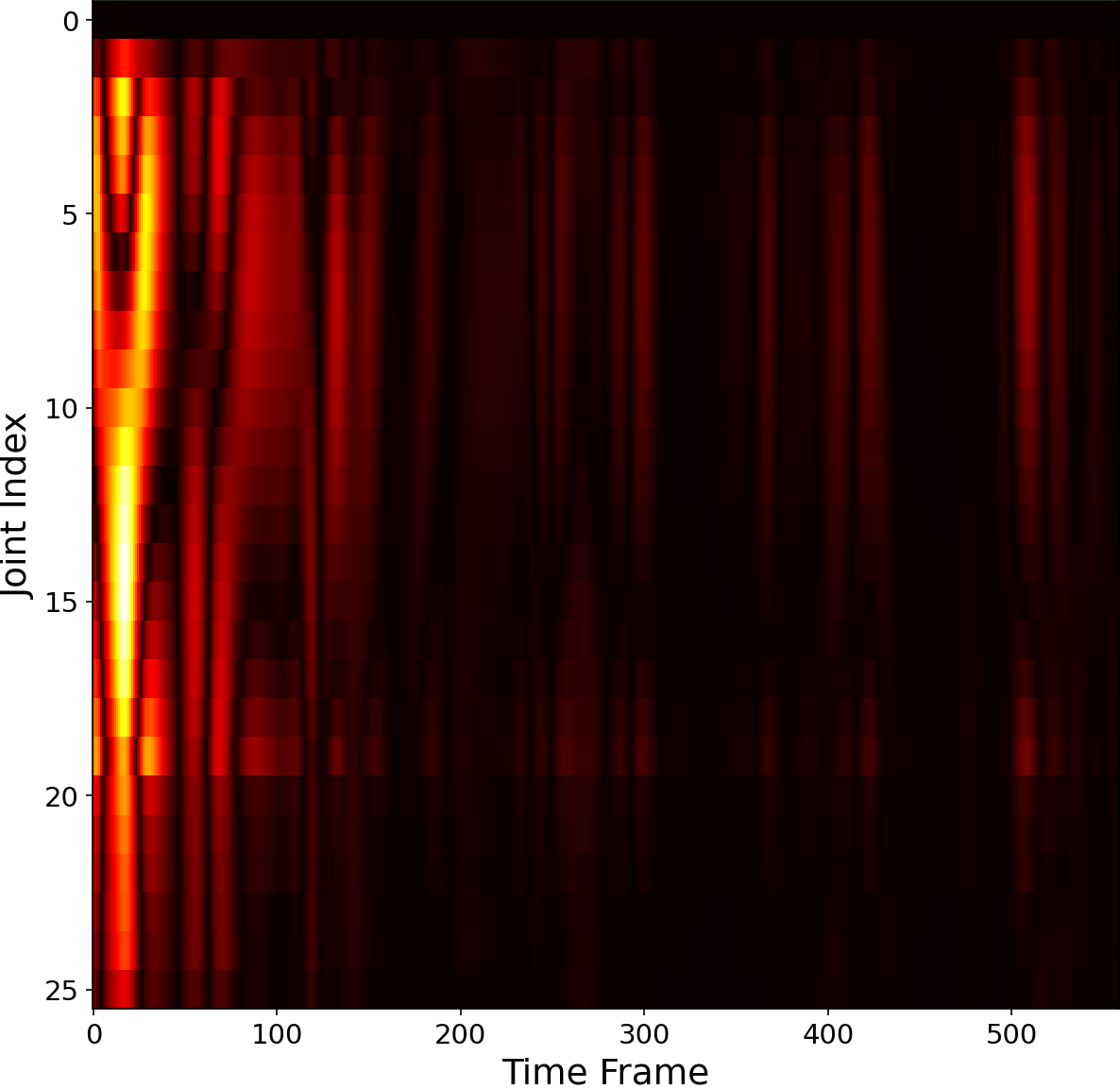}} \\
\end{tabular}
\end{table*}
\section{CMA-ES Fitness Functions}
\label{app:rewards}
We describe the CMA-ES fitness $f(\tau)$ minimized for each task.
$p_{\text{tip}}(t)$ is the rope tip position at simulation time $t$
along trajectory $\tau$, and $p_{\text{tgt}}$ the task target. Numeric
scale factors are fixed hyperparameters tuned once and omitted from
the expressions below for clarity. Candidate trajectories are
pre-filtered at sample time to satisfy joint-velocity limits. Hardware feasibility is also driven into the reward through soft penalties on joint-velocity-limit violations and on collisions, both added to the fitness as positive terms.

\subsection{Shared Quantities}
\begin{align}
d_{\min}        &= \min_t \|p_{\text{tip}}(t) - p_{\text{tgt}}\|_2 \\
R_{\text{prox}} &= \tfrac{1}{1 + d_{\min}}
\end{align}
$M(\tau)$ regularizes total joint motion. $P_{\text{vel}}(\tau) \ge 0$ accumulates the magnitude of joint-velocity-limit violations along $\tau$ and vanishes when all joints stay within limits. $P_{\text{coll}}(\tau) \ge 0$ sums contact-count penalties for rope-robot contacts, extender-pole-robot contacts, and any contact entering the robot's weighted base region, with the base-region term weighted more heavily than the others.

\subsection{Pole Striking}
\begin{equation}
f(\tau) = d_{\min} - R_{\text{prox}} + M(\tau) + P_{\text{vel}}(\tau) + P_{\text{coll}}(\tau)
\end{equation}

\subsection{Lobbing}
The lobbing task augments the striking fitness with a dwell term that
rewards the tip for remaining near the target after arrival. Define
\begin{equation}
T_{\text{near}}(\tau) = \bigl|\{t : \|p_{\text{tip}}(t) - p_{\text{tgt}}\|_2 < \epsilon\}\bigr|,
\end{equation}
the number of simulation timesteps for which the tip is within
distance $\epsilon$ of the target. By construction
$T_{\text{near}}(\tau) > 0$ if and only if $d_{\min} < \epsilon$,
i.e., only when the tip actually enters the target neighborhood at
some point along the trajectory. The dwell reward applies a piecewise
scale to $T_{\text{near}}$:
\begin{equation}
R_{\text{dur}}(\tau) = s(d_{\min}) \cdot T_{\text{near}}(\tau), \qquad
s(d) =
\begin{cases}
\alpha_{\text{hi}} & d < \epsilon \\
\alpha_{\text{lo}} & d \ge \epsilon,
\end{cases}
\end{equation}
with $\alpha_{\text{hi}} \gg \alpha_{\text{lo}}$. The fitness is
\begin{equation}
f(\tau) = d_{\min} - R_{\text{prox}} - R_{\text{dur}} + M(\tau) + P_{\text{vel}}(\tau) + P_{\text{coll}}(\tau).
\end{equation}
This produces a staggered objective. Before any candidate trajectory
achieves arrival, $T_{\text{near}} = 0$, so $R_{\text{dur}}$ vanishes
and the fitness reduces to the pole-striking fitness.
Once a trajectory reaches $d_{\min} < \epsilon$, the scale switches to
$\alpha_{\text{hi}}$ and every additional near-target timestep
contributes a large reduction in fitness. CMA-ES therefore first
finds trajectories that arrive at the target, and then refines them
to maximize dwell time, producing the slow controlled lobbing motion
rather than a fast strike-and-leave.

\subsection{Draping}
Target marked over the wall with an intermediate waypoint over the
wall edge
$p_{\text{wall}} = p_{\text{tgt}} + (\Delta x,\, 0,\, -L/2)$.
Let $d_{\text{tgt}}, d_{\text{wall}}$ be the minimum tip distances to
each, and $P_{\text{height}} = \max\!\bigl(0,\, \max_t p_{\text{tip},z}(t) - z_{\max}\bigr)$
discourage high arcs:
\begin{equation}
f(\tau) = d_{\text{tgt}} + d_{\text{wall}} + P_{\text{height}} {} + P_{\text{vel}}(\tau) + P_{\text{coll}}(\tau).
\end{equation}
$d_{\text{wall}}$ encourages the tip to pass over the wall edge before
descending to the target.

\end{document}